\documentclass[letterpaper]{article} 
\usepackage{aaai2027}  
\usepackage[hyphens]{url}  
\usepackage{graphicx} 
\usepackage{natbib}  
\usepackage{caption} 
\usepackage{algorithm}
\usepackage{amssymb}
\usepackage{enumitem}
\usepackage{tikz}
\usetikzlibrary{arrows}

\usepackage{newfloat}
\usepackage{listings}
\DeclareCaptionStyle{ruled}{labelfont=normalfont,labelsep=colon,strut=off} 
\floatstyle{ruled}
\newfloat{listing}{tb}{lst}{}
\floatname{listing}{Listing}
\usepackage{annotate-equations}
\usepackage{xcolor}
\definecolor{annotpurple}{RGB}{120,60,180}
\definecolor{annotred}{RGB}{180,35,55}
\usepackage{multirow}
\usepackage{cellspace}
\usepackage{algorithm}
\usepackage{algpseudocode} 
\usepackage[noEnd=true]{algpseudocodex}
\usepackage[table]{xcolor}

\definecolor{heatbest}{RGB}{255,166,166}
\definecolor{heatsecond}{RGB}{255,207,166}
\definecolor{heatthird}{RGB}{255,237,166}
\definecolor{heatfourth}{RGB}{255,255,205}
\definecolor{pink}{rgb}{0.929,0.008,0.549}
\usepackage{xcolor}
\definecolor{tocblue}{RGB}{31,73,155}

\newcommand{\tocdots}{\leaders\hbox to 0.55em{\hfil.\hfil}\hfill}
\newcommand{\tocrule}{\par\noindent\rule{\linewidth}{0.6pt}\par}

\newcommand{\tocsec}[3]{%
  \par\vspace{10pt}\noindent
  {\bfseries\color{tocblue}#2\hspace{0.8em}#3}%
  \hfill{\bfseries\pageref{#1}}\par}
\usepackage{booktabs}
\newcommand{\tocsub}[3]{%
  \par\vspace{5pt}\noindent\hspace{1.6em}%
  {\color{tocblue}#2\hspace{0.6em}#3}%
  \hspace{0.4em}\tocdots\hspace{0.4em}\pageref{#1}\par}

\usepackage{amsmath}

\definecolor{cvprblue}{rgb}{0.12,0.49,0.85}

\let\oldcitep\citep
\renewcommand{\citep}[2][]{%
  \textcolor{cvprblue}{\oldcitep[#1]{#2}}%
}
\usepackage{multirow}
\usepackage{graphicx}   
\usepackage{pifont}
\newcommand{\cmark}{\ding{51}}  
\newcommand{\xmark}{\ding{55}}  
\usepackage{arydshln}

\title{TRACE-GS: On-Policy Trajectory Distillation with Privileged
\\Geometric Conditioning for Sparse-View 3DGS Restoration}
\author{
Linlian Jiang\textsuperscript{\rm 1,2},
Yuchen Xi\textsuperscript{\rm 3},
Sadman Rakib Pinon\textsuperscript{\rm 1,2},
Ruigang Yang\textsuperscript{\rm 3},
Yang Wang\textsuperscript{\rm 1,2}\textsuperscript{\textdagger},
Xinxin Zuo\textsuperscript{\rm 1}\textsuperscript{\textdagger}
}
\affiliations{
\textsuperscript{\rm 1}Affiliation 1\\
\textsuperscript{\rm 2}Affiliation 2\\
Email addresses
}
\affiliations{
\vspace{1pt}
    \textsuperscript{\rm 1}Concordia University \quad
    \textsuperscript{\rm 2}Mila -- Quebec AI Institute \quad
    \textsuperscript{\rm 3}Shanghai Jiao Tong University  \\[8pt]
    {\small\tt \{linlian.jiang,sadmanrakib.pinon\}@mail.concordia.ca,} \quad 
    {\small\tt \{yuchen.x,ryang2\}@sjtu.edu.cn,}
    {\small\tt \{yang.wang,xinxin.zuo\}@concordia.ca}\\[10pt]
    Project Page:
    \textcolor{pink}{\url{https://linlianjiang.github.io/trace-gs/}}
    \vspace{3.4pt}

}

\nocopyright
\begin{document}

\maketitle

\begingroup
\renewcommand{\thefootnote}{\textdagger}
\footnotetext{Corresponding authors.}

 \begin{abstract}
\label{sec:abs}


We present \textbf{TRACE-GS}, an on-policy \underline{\textbf{tra}}jectory distillation framework that leverages privileged g\underline{\textbf{e}}ometric \underline{\textbf{c}}onditioning at training time, thereby adapting a diffusion prior to sparse-view 3D Gaussian Splatting (3DGS) restoration. Rather than pursuing increasingly sophisticated restoration architectures, we identify a more fundamental limitation shared by existing diffusion-based approaches: supervision at independently noised states does not cover those reached during inference.
In sparse-view 3DGS, under-constrained geometry biases denoising from
the outset, and the resulting deviations compound along the rollout.
TRACE-GS instead performs on-policy trajectory distillation: a teacher
conditioned on richer geometry from additional training views supplies
targets along the sparse-view student's own rollout, aligning denoising
directions and cross-view responses at each visited state.
This training-only geometry places TRACE-GS in the learning using
privileged information (LUPI) setting.
At deployment, only the sparse-view student is retained, and its restored renderings serve as pseudo-observations for 3DGS refinement.
To the best of our knowledge, TRACE-GS is the first to derive on-policy
supervision from privileged geometry for sparse-view 3DGS restoration,
achieving consistent gains and strong generalization across datasets and
sparse-view settings.





\end{abstract}
\section{Introduction}
\label{sec:intro}

3D reconstruction and novel view synthesis (NVS) are fundamental to
virtual reality~\cite{jiang2024vr}, autonomous
driving~\cite{hess2025splatad}, and robotics~\cite{hsu2026scene}.
For these tasks, 3D Gaussian Splatting
(3DGS)~\cite{kerbl20233d} has emerged as a leading choice for its
real-time rendering.
It achieves high reconstruction quality from dense multi-view
observations, but this quality deteriorates in sparse-view settings
common in practical capture.

Existing methods for sparse-view 3DGS have leveraged various priors to
mitigate ambiguity under limited observations.
One line regularizes the reconstruction through techniques such as
auxiliary depth supervision~\cite{li2024dngaussian,zheng2025nexusgs},
consistency constraints~\cite{zhang2024cor}, and Gaussian
dropout~\cite{park2025dropgaussian}.
While these approaches help reduce overfitting, they lack an explicit
generative mechanism for recovering content in severely under-observed
regions.
A complementary line therefore uses pretrained diffusion models as
generative restoration priors, restoring artifact-laden novel-view
renderings and feeding the outputs back as pseudo-observations for
3DGS refinement~\cite{liu2024deceptive,luo20253denhancer}.
Early work adopts single-image diffusion~\cite{wu2025difix3d+},
producing plausible restorations but processing each view
independently, which can introduce cross-view inconsistencies that
hinder subsequent 3DGS optimization.
This has motivated video diffusion methods~\cite{wu2025genfusion,
yin2025gsfixer}, which jointly model view sequences along camera 
trajectories to enforce consistency across views.

\begin{figure}[t]
\centering
\includegraphics[width=1\linewidth]{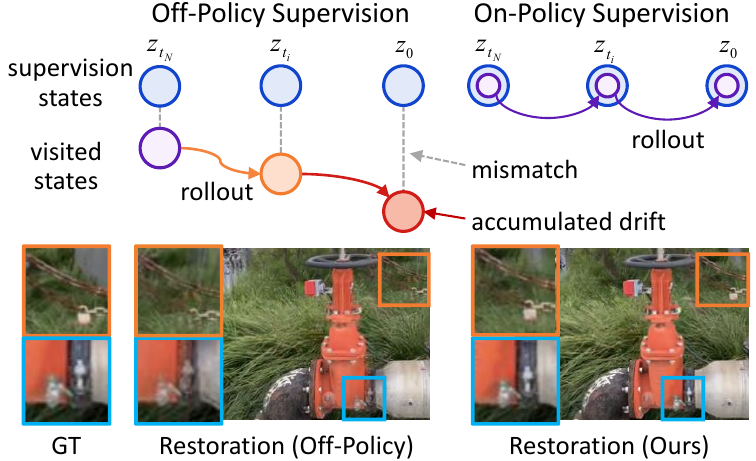}
\caption{Off-policy mismatch in sparse-view 3DGS restoration.
Top: supervision at independently forward-noised states misses rollout
states, letting deviations accumulate; on-policy supervision covers
visited states.
Bottom: on-policy restoration recovers sharper detail from 6 views.}
\label{fig:teaser}
\end{figure}

Although increasingly sophisticated video diffusion architectures
improve sequence-level cross-view modeling, a fundamental
training--inference mismatch remains.
During training, supervision is applied at independently noised states
derived from clean target sequences, whereas inference follows a
sequential rollout driven by the model's own denoising predictions.
Thus, the model encounters self-generated states not directly
covered by the training objective, creating an off-policy state
mismatch~\cite{li2024alleviating} (see Fig.~\ref{fig:teaser}).
In sparse-view 3DGS restoration, this mismatch interacts with
under-constrained geometry: artifact-laden renderings bias denoising
directions from the outset, and the resulting deviations can compound
along the rollout~\cite{ning2024elucidating}.
To address these coupled challenges, we seek a training signal that
both \textbf{(i)} provides reliable denoising directions and
\textbf{(ii)} supervises the model at states it actually encounters
along its inference trajectory.
On-policy distillation offers a natural framework for the latter by
supervising the model along its own rollout, and has shown promise in
distilling autoregressive
models~\cite{agarwal2024policy,gu2024minillm}.
Its effectiveness, however, depends critically on the availability of reliable guidance at these on-policy states. 
We observe that training scenes typically contain additional views from which a better-constrained 3DGS can be reconstructed, yielding renderings with more reliable geometry and fewer artifacts. 
When conditioned on these dense-view renderings, the same restoration
model can therefore provide less biased denoising directions without requiring
architectural changes.
Because the geometry enabled by these additional views is available
only during training, it constitutes privileged information in the
learning using privileged information (LUPI)
setting~\cite{vapnik2009new,lopez2015unifying}: it can guide model
learning, while deployment relies solely on sparse-view inputs.

We thus introduce TRACE-GS, an on-policy trajectory distillation framework built on a coherent supervision mechanism: the teacher uses privileged geometry to provide reliable denoising targets, while the student's own rollout determines where they are applied.
By design, teacher and student have identical architecture and
capacity, with their conditioning asymmetry arising solely from the
geometry of dense- and sparse-view 3DGS renderings.
We first align their predictions at independently noised states, 
reducing the student's initial direction bias.
Unlike prior diffusion-based refinement methods~\cite{wu2025genfusion,yin2025gsfixer}, 
whose supervision remains confined to such independently sampled states, 
TRACE-GS extends supervision to the rollout states themselves, querying the frozen
teacher at each one without separately propagating teacher states.
To further constrain how the student uses cross-view evidence beyond
what direction matching captures, we introduce on-policy retrieval
alignment that matches the two paths' retrieval responses along that
trajectory.
At deployment, only the sparse-conditioned student is retained; its
restored renderings serve as pseudo-observations for sparse-view 3DGS
refinement.

Our contributions can be summarized as follows:
\begin{itemize}
\item We identify that off-policy mismatch and geometry-induced bias
compound in sparse-view 3DGS restoration, and address both by coupling
reliable targets with student-visited states.
\item We formulate geometry-asymmetric LUPI, deriving velocity and
cross-view targets from an identical-capacity teacher with training-only
geometric conditioning.
\item To the best of our knowledge, TRACE-GS is the first to supervise
sparse-view 3DGS restoration on-policy, using privileged dense-view
geometry to define the target at each visited state. Experiments across
datasets and sparsity levels show strong performance and generalization.
\end{itemize}
\section{Related Work}
\label{sec:related_work}

\noindent\textbf{Sparse-View 3DGS Restoration with Diffusion Priors.}
Sparse-view 3DGS is severely under-constrained, often producing
floaters, fragmented geometry, and blurred textures in under-observed
regions~\cite{wu2025genfusion,yin2025gsfixer}.
Regularization-based methods mitigate these failures using auxiliary
depth supervision~\cite{li2024dngaussian,zheng2025nexusgs},
consistency constraints~\cite{zhang2024cor}, or Gaussian
dropout~\cite{park2025dropgaussian}.
Diffusion priors have also been incorporated at the reconstruction
level: ReconFusion~\cite{wu2024reconfusion} regularizes NeRF
optimization, whereas OracleGS~\cite{topalouglu2026oraclegs} weights
diffusion-generated views by MVS uncertainty during 3DGS optimization.
A complementary line restores artifact-laden novel-view renderings with
pretrained diffusion models and feeds them back as pseudo-observations
for 3DGS refinement.
Difix3D+~\cite{wu2025difix3d+} uses single-image diffusion, whereas
GenFusion and GSFixer~\cite{wu2025genfusion,yin2025gsfixer} jointly
restore view sequences with video diffusion.
Despite different view-coupling strategies, these restorers supervise
independently forward-noised states rather than those reached by the
sparse-conditioned rollout.
This leaves conditioning-induced errors uncorrected along the rollout,
limiting the restoration prior's effectiveness.

\noindent\textbf{On-Policy Distillation.}
Diffusion distillation methods~\cite{yin2024one,luo2023diff,%
zhou2024score} primarily target one-step generation, supervising states
obtained by noising student outputs rather than intermediate states
along a multi-step rollout.
For multi-step diffusion, rollout states remain unsupervised, creating
a training--inference mismatch linked to exposure
bias~\cite{ning2023input,li2024alleviating,ning2024elucidating}.
On-policy distillation supervises states along student rollouts, following on-policy imitation learning~\cite{ross2011reduction};
related formulations show promise in autoregressive
LLMs~\cite{agarwal2024policy,gu2024minillm,zhao2026self}, few-step
diffusion trajectory matching~\cite{luo2025learning}, and diffusion
self-correction~\cite{qin2026soar}.
On-policy supervision remains unexplored in sparse-view 3DGS
restoration, where reliable targets are unavailable from sparse
conditioning alone; TRACE-GS derives them from training-time dense-view
geometry.

\noindent\textbf{Learning Using Privileged Information.}
Learning using privileged information (LUPI) exploits information
available during training but absent at deployment
~\cite{vapnik2009new}.
Generalized distillation transfers knowledge derived from such
information through a privileged teacher to a deployable
student~\cite{lopez2015unifying}, with recent extensions addressing
missing-modality settings~\cite{chen2024probabilistic}.
Recent deep formulations use privileged information to model noisy
supervision~\cite{collier2022transfer}, transfer knowledge from
enhanced training inputs~\cite{tzortzis2024learning}, or guide policies
with privileged task states~\cite{messikommer2025student}.
However, access alone does not guarantee
transfer~\cite{provodin2024rethinking}, motivating careful design of
how the privileged signal is transferred to the deployable model.
Unlike these formulations, TRACE-GS derives its privileged signal from
the geometric fidelity gap between dense- and sparse-view renderings,
yielding a teacher--student pair that differs only in geometric
conditioning.

\begin{figure*}[t] 
\centering \includegraphics[width=1\linewidth]{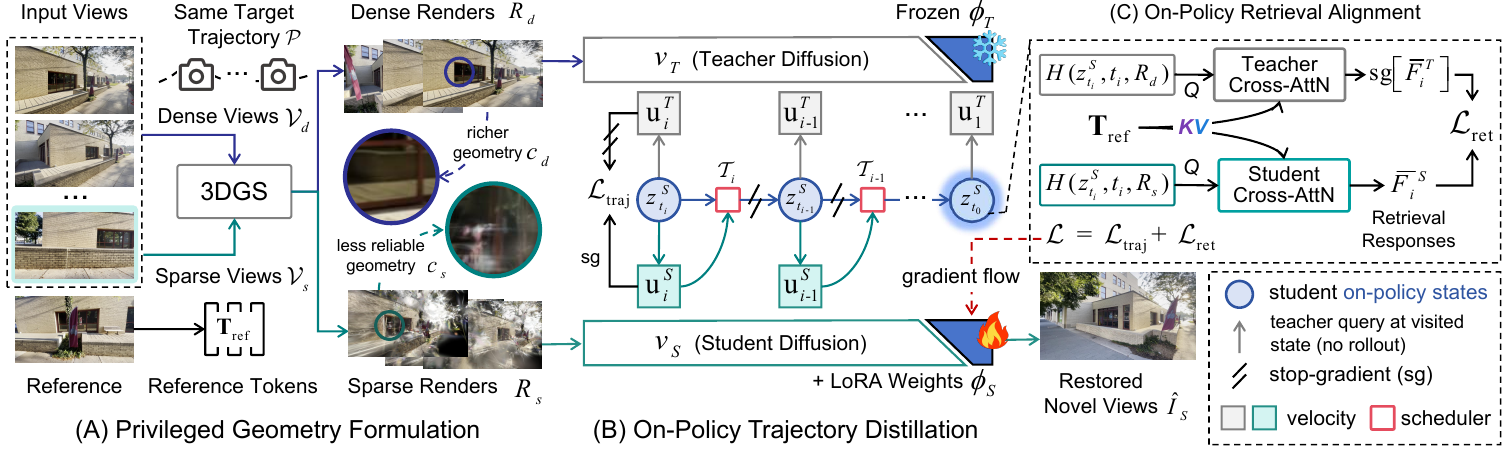} 
\caption{
Overview of TRACE-GS.
\textbf{(A)} Given $\mathcal V_s\subset\mathcal V_d$, we independently
reconstruct a 3DGS from each view set and render paired sequences
$R_s$ and $R_d$ along the same target trajectory.
\textbf{(B)} Built on the same frozen video-diffusion backbone, the
teacher and student use a frozen LoRA $\phi_T$ and a trainable LoRA
$\phi_S$ cloned from a shared initialization, and are conditioned on
$R_d$ and $R_s$, respectively, forming a geometry-asymmetric pair.
The student rolls out its reverse trajectory, while the teacher is
queried at the same student-visited states without its own rollout.
\textbf{(C)} At these states, $\mathcal L_{\mathrm{traj}}$ aligns
velocity predictions, while $\mathcal L_{\mathrm{ret}}$ aligns
cross-view retrieval responses.
Only $\phi_S$ is updated and retained for deployment.
}
\label{fig:structure} 
\end{figure*}

\section{Method}
\label{sec:method}

We present TRACE-GS, an on-policy trajectory distillation framework
using training-time privileged geometry to supervise student-visited states
in sparse-view 3DGS restoration.
Fig.~\ref{fig:structure}
illustrates the geometry-asymmetric LUPI formulation
(Sec.~\ref{sec:motivation}) and on-policy distillation along the
student rollout (Sec.~\ref{sec:trajectory}); the preceding
direction alignment is described in Sec.~\ref{sec:supervision}.
Once trained, only the student model is deployed.

\subsection{Preliminary}
\noindent\textbf{Sparse-View 3D Gaussian Splatting.}
3D Gaussian Splatting (3DGS)~\cite{kerbl20233d} reconstructs a scene
from $K$ posed images
$\mathcal{V}_s=\{(I_i,P_i)\}_{i=1}^{K}$ (each image $I_i$ paired with
camera pose $P_i$) as a set of learnable Gaussian primitives
$\mathcal{G}$, which are rendered to novel views via differentiable
alpha compositing. We focus on the sparse-view
setting~\cite{wu2025genfusion}, where $K$ is small
(e.g., $K\in\{3,6,9\}$). Given a target camera trajectory
$\mathcal{P}=\{P_n\}_{n=1}^{L}$, the resulting novel-view sequence is
$R=\{\tilde I_n=\operatorname{Render}(\mathcal{G},P_n)\}_{n=1}^{L}$.

\noindent\textbf{Diffusion-Based Restoration.}
Diffusion-based restoration adapts a video diffusion
model~\cite{yang2024cogvideox,wan2025wan} to restore an artifact-laden
sequence $R$ from a sparse-view 3DGS $\mathcal G$.
Given a denoising schedule $t_M>\cdots>t_0$, it starts from
$z_{t_M}\sim\mathcal{N}(0,\mathbf I)$ and updates
\[
z_{t_{i-1}}
=
\mathcal{T}_i\!\left(
z_{t_i},
v_\theta(z_{t_i},t_i,R)
\right),
\]
where $\mathcal{T}_i$ is the scheduler update to $t_{i-1}$.
The decoded sequence
$\widehat I=\{\widehat I_n\}_{n=1}^{L}$ is paired with its target poses
$\{P_n\}_{n=1}^{L}$.
During deployment, these posed pseudo-observations and $\mathcal V_s$
supervise iterative optimization of $\mathcal G$, yielding its refined
version $\mathcal G'$ for final novel-view inference.

\subsection{Geometry-Asymmetric Learning with Privileged Information}
\label{sec:motivation}

\noindent\textbf{Motivation.}
Reliable guidance is difficult to obtain from sparse-view conditioning
alone.
At deployment, sparse-view restoration
methods~\cite{wu2025genfusion,yin2025gsfixer} operate with only a few
input views (e.g., 3, 6, or 9), typically resulting in incomplete
geometry and artifact-prone renderings.
Training scenes often provide many more views. Yet, to match the
sparse deployment setting, existing methods restrict
conditioning to the same sparse subset, leaving the remaining views
and their geometric evidence unexploited.

\begin{center}
\emph{
The natural question is: how can we transfer this training-time geometric advantage to a restorer \\that sees only sparse views at deployment?
}

\end{center}
Knowledge distillation~\cite{hinton2015distilling} offers a principled
mechanism for transferring this advantage and is commonly instantiated
through a teacher--student architecture. 
Unlike conventional formulations~\cite{sengupta2024good}, which
typically rely on \textbf{\textit{model-capacity asymmetry}} to
transfer knowledge from a high-capacity teacher to a smaller student,
we introduce a new formulation for sparse-view 3DGS restoration, which
we call \textbf{\textit{geometry asymmetry}}: 
knowledge is transferred from a teacher whose conditioning carries
more reliable geometry to an architecturally identical, equal-capacity
student conditioned on less reliable geometry.

\noindent\textbf{Privileged Geometry Formulation.}
Let $\mathcal V_s\subset\mathcal V_d$ denote the sparse and dense view
sets.
As shown in Fig.~\ref{fig:structure}(A), we reconstruct a 3DGS from
each and render both along the same target trajectory $\mathcal P$,
yielding paired sequences $R_s$ and $R_d$.
The restoration model maps $R_d$ to a training-only pseudo-target
$\widehat I_T$ and $R_s$ to deployable pseudo-observations
$\widehat I_S$:
\begin{equation}
\begin{aligned}
\mathcal{V}_d
&\xrightarrow{\;\mathrm{3DGS}\;}
R_d
\overset{\scriptscriptstyle\mathrm{encodes}}{\dashrightarrow}
c_d
\xrightarrow{\;\mathrm{restore}\;}
\widehat{I}_T
\qquad \qquad\text{\small(training only)},\\
\mathcal{V}_s
&\xrightarrow{\;\mathrm{3DGS}\;}
R_s
\overset{\scriptscriptstyle\mathrm{encodes}}{\dashrightarrow}
c_s
\xrightarrow{\;\mathrm{restore}\;}
\widehat{I}_S
\overset{\mathrm{update}}{\rightsquigarrow}
\mathcal{G}'
\;\text{\small(deployment)} .
\end{aligned}
\label{eq:conditioning}
\end{equation}
Here, $c_b$ denotes the geometry implicitly carried by $R_b$ for
$b\in\{s,d\}$, with dashed arrows marking this association.
Since $R_d$ is rendered from a 3DGS fitted with more views, it typically
carries more reliable geometry than $R_s$.

Because the additional views
$\mathcal V_d\setminus\mathcal V_s$ are available during training but
not at deployment, the richer geometry $c_d$ they enable constitutes
\emph{privileged information}. This naturally casts sparse-view
restoration as a \textbf{\emph{learning using privileged information}
(LUPI)} problem~\cite{vapnik2009new,lopez2015unifying}.
Accordingly, the dense-view path provides privileged training
supervision, whereas only the sparse-view output $\widehat I_S$
updates the 3DGS at deployment.

\subsection{Privileged Geometry Supervision}
\label{sec:supervision}

\noindent\textbf{Geometry-Asymmetric Teacher--Student Roles.}
To instantiate the geometry-asymmetric LUPI formulation, we define a
privileged teacher role and a deployable student role within a single
velocity predictor
$v_{\theta,\phi}$~\cite{yang2024cogvideox}.
The two roles share the frozen backbone $\theta$ and a trainable
LoRA $\phi$, differing only in their conditioning.
They receive the same $z_t$, obtained by forward-noising the latent
encoding of $\widehat I_T$ with a shared noise sample $\epsilon$ at
timestep $t$, and are conditioned on $R_d$ and $R_s$, respectively,
which correspond to the same target-view sequence:
\begin{equation}
\underbrace{
v_T = v_{\theta,\phi}(z_t,\, t,\, R_d)
}_{\substack{\text{teacher role (privileged)} \\ R_d \dashrightarrow c_d,\; |\mathcal V_d|>|\mathcal V_s|}},
\qquad
\underbrace{
v_S = v_{\theta,\phi}(z_t,\, t,\, R_s)
}_{\substack{\text{student role (standard)} \\ R_s \dashrightarrow c_s,\; |\mathcal{V}_s| = K \in \{3,6,9\}}}.
\label{eq:teacher_student}
\end{equation}
The dense-conditioned prediction $v_T$ therefore provides the
privileged supervisory direction for $v_S$.

\noindent\textbf{Privileged Direction Alignment.}
To reduce the direction gap induced by the different geometric
reliability of $R_s$ and $R_d$, we align the two paths' single-step
predictions at forward-noised states:
\begin{equation}
\mathcal L_{\mathrm{align}}
=
\mathbb E_{(R_s,R_d),\,t,\,\epsilon}
\left[
\left\|
v_S-\mathrm{sg}\!\left[v_T\right]
\right\|_2^2
\right],
\label{eq:align}
\end{equation}
where $\epsilon\sim\mathcal N(0,\mathbf I)$ and $\mathrm{sg}[\cdot]$
denotes stop-gradient, so the alignment is directional: it pulls $v_S$
toward the dense-conditioned direction without backpropagating through
$v_T$.
Both paths receive standard flow-matching supervision on
$\widehat I_T$, yielding
$\mathcal L_{\mathrm{warm}}=\mathcal L_{\mathrm{fm}}^{S}
+\mathcal L_{\mathrm{fm}}^{T}
+\lambda_{\mathrm{align}}\mathcal L_{\mathrm{align}}$,
anchoring both predictions to the pseudo-target velocity and
discouraging a conditioning-invariant shortcut.
Because $\phi$ is shared, $v_T$ is recomputed each update and
remains an online target.


\vspace{-2pt}

\subsection{On-Policy Trajectory Distillation with Privileged Geometry}
\label{sec:trajectory}
The privileged direction alignment in Sec.~\ref{sec:supervision}
provides one-step dense-conditioned supervision only at independently
sampled forward-noised states, leaving the recursive student rollout
unaddressed and creating two coupled limitations:
\begin{enumerate}[label=(\roman*),leftmargin=*,itemsep=2pt]
\item \textbf{Student-visited states lack direct supervision.}
This omission creates an off-policy gap analogous to that in imitation
learning~\cite{ross2011reduction}.

\item \textbf{Geometry-induced errors compound along the rollout.}
Under-constrained sparse-view geometry biases denoising directions from
the outset; as each state builds on the previous prediction, the
resulting errors accumulate~\cite{ning2023input,li2024alleviating}.
\end{enumerate}
These effects interact: biased directions push the student into
uncovered states, where subsequent predictions can deviate further.
We therefore query the dense-conditioned teacher at each visited state, 
as shown in Fig.~\ref{fig:structure}(B).

\noindent\textbf{Student On-Policy Rollout.}
With $\phi_T$ frozen, the student rolls out on a time grid
$\{t_i'\}_{i=0}^{N}$ resampled each iteration from the deployment
schedule. Starting from
$z_{t_N'}^S\sim\mathcal N(0,\mathbf I)$, it updates for
$i=N,\ldots,1$:
\begin{equation}
u_i^S=v_{\theta,\phi_S}(z_{t_i'}^S,t_i',R_s),
\quad
z_{t_{i-1}'}^S=
\mathrm{sg}\!\left[
\mathcal T_i(z_{t_i'}^S,u_i^S)
\right].
\label{eq:rollout}
\end{equation}
The visited trajectory is
$\mathcal Z_S=\{(z_{t_i'}^S,t_i')\}_{i=1}^{N}$.

\noindent\textbf{Privileged Teacher Supervision.}
At each student-visited state, we query the frozen teacher without a
separate rollout, so supervision stays anchored to the states actually
visited:
\begin{equation}
u_i^T = v_{\theta,\phi_T}(z_{t_i'}^S,t_i',R_d),\quad i=1,\ldots,N.
\label{eq:teacher_velocity}
\end{equation}
This supplies privileged supervision at states not covered by the
preceding direction alignment.

\noindent\textbf{Trajectory Distillation Loss.}
We align the student predictions with the frozen teacher targets along
$\mathcal Z_S$:
\begin{equation}
\mathcal L_{\mathrm{traj}}
=
\mathbb E_{(R_s,R_d),\,z_{t_N'}^S}
\left[
\frac{1}{N}\sum_{i=1}^{N}
\left\|
u_i^S-\mathrm{sg}[u_i^T]
\right\|_2^2
\right].
\label{eq:traj_loss}
\end{equation}
Gradients flow to $\phi_S$ through $u_i^S$ at each visited state, but
not through the detached scheduler transitions.

\noindent\textbf{On-Policy Retrieval Alignment.}
The teacher and student paths attend to the same timestep-invariant
reference-view tokens $\mathbf T_{\mathrm{ref}}$~\cite{yin2025gsfixer},
but path-specific LoRA projections can yield different retrieval
responses.
Velocity matching aligns denoising directions but does not directly
constrain these responses.
We therefore introduce on-policy retrieval alignment to match the
student's responses to the teacher's at the same visited states, as
illustrated in Fig.~\ref{fig:structure}(C).
\vspace{17pt}
\begin{equation}
F_i^{b}
=
\operatorname{Softmax}\!\left(
  \eqnmark[teal]{query}{H_i^{b}}W_Q^{b}
  \left(
    \eqnmark[annotpurple]{key}{\mathbf{E}^{b}}
  \right)^{\!\top}
  /\sqrt{d}
\right)
\left(\eqnmark[cvprblue]{memory}{\mathbf{T}_{\mathrm{ref}}}\textcolor{cvprblue}{W_V^{b}}\right),
\label{eq:retrieval}
\end{equation}
\annotate[yshift=-0.6em]{below,right}{query}{%
  $H(z_{t_i'}^S,t_i',R_b;\theta,\phi_b),\ b\in\{S,T\}$}
\annotate[yshift=0.7em]{above,left}{key}{%
  $\mathbf{E}^{b}=
 {\mathbf{T}_{\mathrm{ref}}}W_K^{b}$,
  reference keys}
\annotate[yshift=0.7em]{above,right}{memory}{%
  ref. values}
\vspace{2pt}

\noindent where $b\in\{S,T\}$ indexes paths conditioned on $R_s$ and
$R_d$, respectively; $H_i^b$ are hidden states, and
$W_{Q,K,V}^b$ include $\phi_b$.

\begin{algorithm}[t]
\caption{On-Policy Distillation and Deployment}
\label{alg:trace}
\begin{algorithmic}[1]
\Require $\mathcal V_s,\mathcal V_d,\mathcal P$;
deployment schedule $\{t_j\}_{j=0}^{M}$, solver $\{\mathcal T_i\}$;
$N,N_r,\lambda_{\mathrm{ret}}$; frozen $\theta$; aligned $\phi$
(Sec.~\ref{sec:supervision})
\State $\phi_T\leftarrow\mathrm{Freeze}(\mathrm{Clone}(\phi))$,\;
       $\phi_S\leftarrow\mathrm{Clone}(\phi)$
\State $R_b\leftarrow
\mathrm{Render}(\mathrm{3DGS}(\mathcal V_b),\mathcal P)$,
$b\in\{s,d\}$

\Statex {\textbf{// \textcolor{annotred}{Training}: On-policy distillation}}
\While{not converged}
    \State $\{t_i'\}_{i=0}^{N}\leftarrow
    \mathrm{Resample}(\{t_j\}_{j=0}^{M},N)$
    \Comment{training grid}
    \State $z_{t_N'}^S\sim\mathcal N(\mathbf0,\mathbf I)$
    \Comment{initial noise}
    \For{$i=N,\ldots,1$}
        \State $(u_i^b,F_i^b)\leftarrow
        v_{\theta,\phi_b}(z_{t_i'}^S,t_i',R_b)$,
        $b\in\{S,T\}$
        \Comment{student/teacher}
        \State $z_{t_{i-1}'}^S\leftarrow
        \mathrm{sg}[\mathcal T_i(z_{t_i'}^S,u_i^S)]$
        \Comment{detached transition}
    \EndFor
    \State Update $\phi_S$ with
    $\mathcal L_{\mathrm{traj}}
    +\lambda_{\mathrm{ret}}\mathcal L_{\mathrm{ret}}$
\EndWhile

\Statex {\textbf{// \textcolor{cvprblue}{Deployment}: 3DGS refinement}}
\State $\mathcal G\leftarrow\mathrm{3DGS}(\mathcal V_s)$
\For{$r=1,\ldots,N_r$}
    \State $\widehat I_S\leftarrow
    \mathrm{Restore}(\theta,\phi_S,\mathrm{Render}(\mathcal G,\mathcal P))$
    \State $\mathcal G\leftarrow
    \mathrm{Update}(\mathcal G;\mathcal V_s,\widehat I_S,\mathcal P)$
    \Comment{pseudo-obs.\ update}
\EndFor
\State $\mathcal G'\leftarrow\mathcal G$
\end{algorithmic}
\end{algorithm}

\noindent Let $\bar F_i^b=\operatorname{Norm}(F_i^b)$ denote the token-wise
$\ell_2$-normalized response. We align the student response with its
frozen privileged counterpart:
\begin{equation}
\mathcal L_{\mathrm{ret}}
=
\mathbb E_{(R_s,R_d),\,z_{t_N'}^S}
\left[
\frac{1}{N}
\sum_{i=1}^{N}
\left\|
\bar F_i^S-\mathrm{sg}\!\left[\bar F_i^T\right]
\right\|_2^2
\right].
\label{eq:feat_loss}
\end{equation}
The overall on-policy training objective is
\begin{equation}
\mathcal L
=
\mathcal L_{\mathrm{traj}}
+
\lambda_{\mathrm{ret}}\mathcal L_{\mathrm{ret}}.
\label{eq:total_loss}
\end{equation}

\noindent\textbf{Deployment-Time 3DGS Refinement.}
After training, only the student $\phi_S$ is retained. Given a scene
reconstructed from $K$ sparse views, the student restores its
artifact-prone renderings $R_s$ without access to the dense-view
reconstruction or the teacher. The restored renderings serve as
pseudo-observations for refining the 3DGS representation $\mathcal G$
(cf.\ Eq.~\eqref{eq:conditioning}), alternating with 3DGS optimization
for a fixed number of rounds.
Alg.~\ref{alg:trace} summarizes the full procedure.

\section{Experiments}
\label{sec:exp}

\subsection{Experimental Settings}

\noindent\textbf{Implementation Details.}
We freeze a pretrained video-diffusion restoration
backbone~\cite{yin2025gsfixer,yang2024cogvideox} and train rank-32 LoRA
adapters at $480\times720$ resolution.
Training uses AdamW with a learning rate of $1\times10^{-5}$, batch size
$2$/GPU, $\lambda_{\mathrm{align}}=1$, and
$\lambda_{\mathrm{ret}}=0.15$.
Each iteration resamples an $N=10$-step rollout from the $M=50$-step
deployment schedule.
Deployment alternates restoration and 3DGS optimization for $N_r=3$
rounds on one NVIDIA A100 GPU (see the supplement for computational
costs).


\noindent\textbf{Training Data.}
We construct the training set from 112 scenes of
DL3DV~\cite{ling2024dl3dv}.
For each scene, we sample $K\in\{3,6,9\}$ input views to reconstruct
a sparse-view 3DGS and use all available training views for its
dense-view counterpart.
We render both along shared trajectories, obtaining 150 paired clips
per scene.
Sparse-view renderings from 7K and 17K iterations form the student
input $R_s$ and are paired with the same dense-view renderings
$R_d$ from 30K iterations.
A frozen pretrained restorer generates pseudo-targets $\widehat I_T$
from $R_d$. The latent encodings of $\widehat I_T$ are forward-noised
to sample the warm-up states.
Each clip contains 49 frames, and both paths share the same pair of
reference views.
\begin{table}[t]
\centering
\begingroup
\setlength{\tabcolsep}{4pt}
\renewcommand{\arraystretch}{1.05}
\resizebox{\columnwidth}{!}{%
\begin{tabular}{lccc|ccc|ccc} 
\toprule
& \multicolumn{3}{c|}{PSNR $\uparrow$}
& \multicolumn{3}{c|}{SSIM $\uparrow$}
& \multicolumn{3}{c}{LPIPS $\downarrow$} \\
\cmidrule(lr){2-4}\cmidrule(lr){5-7}\cmidrule(lr){8-10}
Method
& 3 & 6 & 9
& 3 & 6 & 9
& 3 & 6 & 9 \\
\midrule

3DGS
& 13.72 & 17.11 & 19.05
& 0.410 & 0.547 & 0.625
& 0.521 & 0.372 & 0.293 \\

Difix3D+
& \cellcolor{heatthird}15.07
& \cellcolor{heatfourth}18.26
& \cellcolor{heatfourth}20.12
& \cellcolor{heatfourth}0.481
& \cellcolor{heatfourth}0.589
& \cellcolor{heatfourth}0.656
& \cellcolor{heatsecond}0.473
& \cellcolor{heatbest}0.329
& \cellcolor{heatbest}0.259 \\

GenFusion
& \cellcolor{heatfourth}14.64
& \cellcolor{heatthird}18.36
& \cellcolor{heatthird}20.32
& \cellcolor{heatthird}0.498
& \cellcolor{heatthird}0.610
& \cellcolor{heatbest}0.688
& \cellcolor{heatfourth}0.493
& \cellcolor{heatfourth}0.374
& \cellcolor{heatfourth}0.314 \\

GSFixer
& \cellcolor{heatsecond}16.21
& \cellcolor{heatsecond}19.11
& \cellcolor{heatsecond}20.60
& \cellcolor{heatsecond}0.536
& \cellcolor{heatsecond}0.625
& \cellcolor{heatthird}0.675
& \cellcolor{heatthird}0.478
& \cellcolor{heatthird}0.360
& \cellcolor{heatthird}0.307 \\

Ours
& \cellcolor{heatbest}16.92
& \cellcolor{heatbest}19.46
& \cellcolor{heatbest}21.07
& \cellcolor{heatbest}0.553
& \cellcolor{heatbest}0.639
& \cellcolor{heatsecond}0.680
& \cellcolor{heatbest}0.463
& \cellcolor{heatsecond}0.349
& \cellcolor{heatsecond}0.295 \\

\bottomrule
\end{tabular}%
}
\endgroup
\caption{Quantitative comparison on DL3DV-Benchmark with 3, 6, and 9 input views. Colors from
{\setlength{\fboxsep}{1pt}\colorbox{heatfourth}{yellow}} to
{\setlength{\fboxsep}{1pt}\colorbox{heatbest}{red}}
indicate increasing performance.}
\label{tab:dl3dv}
\end{table}

\begin{table*}[t]
\centering
\setlength{\tabcolsep}{4.3pt}
\renewcommand{\arraystretch}{1.1}
\resizebox{0.995\textwidth}{!}{%
\begin{tabular}{lccc|ccc|ccc|ccc|ccc|ccc}
\toprule
            & \multicolumn{9}{c|}{\textbf{Mip-NeRF 360}} & \multicolumn{9}{c}{\textbf{NeRFBusters}} \\
\cmidrule(lr){2-10}\cmidrule(lr){11-19}
            & \multicolumn{3}{c|}{PSNR $\uparrow$} & \multicolumn{3}{c|}{SSIM $\uparrow$} & \multicolumn{3}{c|}{LPIPS $\downarrow$}
            & \multicolumn{3}{c|}{PSNR $\uparrow$} & \multicolumn{3}{c|}{SSIM $\uparrow$} & \multicolumn{3}{c}{LPIPS $\downarrow$} \\
\cmidrule(lr){2-4}\cmidrule(lr){5-7}\cmidrule(lr){8-10}
\cmidrule(lr){11-13}\cmidrule(lr){14-16}\cmidrule(lr){17-19}
Method      & 3 & 6 & 9 & 3 & 6 & 9 & 3 & 6 & 9 & 3 & 6 & 9 & 3 & 6 & 9 & 3 & 6 & 9 \\
\midrule
Zip-NeRF (ICCV'23)   & 12.77 & 13.61 & 14.30 & 0.271 & 0.284 & 0.312 & 0.705 & 0.663 & 0.633 & 13.00 & 13.50 & 14.09 & 0.354 & 0.369 & 0.380 & 0.653 & 0.620 & 0.580 \\
ReconFusion$^{*}$ (CVPR'24) & 15.50 & 16.93 & 18.19 & 0.358 & 0.401 & 0.432 & 0.585 & 0.544 & 0.511 & 15.62  & 17.12 & 18.26  & 0.371  & 0.415   & 0.439 & 0.574  & 0.558  & 0.527  \\
\midrule
3DGS (SIGGRAPH'23)        & 13.06 & 14.96 & 16.79 & 0.251 & 0.355 & 0.447 & 0.576 & 0.505 & 0.446 & 13.84 & 15.98 & 18.12 & 0.415 & 0.502 & 0.574 & 0.587 & 0.503 & 0.441 \\
2DGS (SIGGRAPH'24)       & 13.07 & 15.02 & 16.67 & 0.243 & 0.338 & 0.423 & 0.580 & 0.506 & 0.449 & 13.18 & 15.56 & 17.81 & 0.387 & 0.480 & 0.571 & 0.522 & 0.437 & 0.358 \\
Difix3D+ (CVPR'25)
& \cellcolor{heatfourth}13.92
& \cellcolor{heatfourth}15.94
& \cellcolor{heatfourth}17.54
& \cellcolor{heatfourth}0.298
& \cellcolor{heatfourth}0.382
& \cellcolor{heatfourth}0.452
& \cellcolor{heatthird}0.578
& \cellcolor{heatsecond}0.468
& \cellcolor{heatbest}0.391
& \cellcolor{heatthird}15.48
& \cellcolor{heatthird}18.46
& \cellcolor{heatfourth}20.05
& \cellcolor{heatfourth}0.457
& \cellcolor{heatfourth}0.581
& \cellcolor{heatfourth}0.636
& \cellcolor{heatsecond}0.473
& \cellcolor{heatbest}0.354
& \cellcolor{heatbest}0.302 \\
GenFusion (CVPR'25)
& \cellcolor{heatthird}15.03
& \cellcolor{heatthird}16.90
& \cellcolor{heatthird}18.29
& \cellcolor{heatthird}0.357
& \cellcolor{heatsecond}0.430
& \cellcolor{heatsecond}0.489
& \cellcolor{heatthird}0.578
& \cellcolor{heatfourth}0.494
& \cellcolor{heatfourth}0.440
& \cellcolor{heatfourth}14.93
& \cellcolor{heatfourth}18.05
& \cellcolor{heatsecond}20.28
& \cellcolor{heatthird}0.516
& \cellcolor{heatsecond}0.601
& \cellcolor{heatsecond}0.669
& \cellcolor{heatfourth}0.505
& \cellcolor{heatfourth}0.396
& \cellcolor{heatthird}0.324 \\
GSFixer (ICML'26)
& \cellcolor{heatsecond}15.61
& \cellcolor{heatsecond}17.27
& \cellcolor{heatsecond}18.63
& \cellcolor{heatsecond}0.370
& \cellcolor{heatthird}0.426
& \cellcolor{heatthird}0.481
& \cellcolor{heatsecond}0.559
& \cellcolor{heatthird}0.478
& \cellcolor{heatthird}0.420
& \cellcolor{heatsecond}16.00
& \cellcolor{heatsecond}18.54
& \cellcolor{heatsecond}20.28
& \cellcolor{heatsecond}0.519
& \cellcolor{heatthird}0.597
& \cellcolor{heatthird}0.651
& \cellcolor{heatthird}0.482
& \cellcolor{heatthird}0.395
& \cellcolor{heatfourth}0.335 \\
Ours
& \cellcolor{heatbest}16.34
& \cellcolor{heatbest}17.99
& \cellcolor{heatbest}19.20
& \cellcolor{heatbest}0.396
& \cellcolor{heatbest}0.455
& \cellcolor{heatbest}0.506
& \cellcolor{heatbest}0.543
& \cellcolor{heatbest}0.450
& \cellcolor{heatsecond}0.402
& \cellcolor{heatbest}16.60
& \cellcolor{heatbest}19.26
& \cellcolor{heatbest}20.79
& \cellcolor{heatbest}0.540
& \cellcolor{heatbest}0.619
& \cellcolor{heatbest}0.678
& \cellcolor{heatbest}0.467
& \cellcolor{heatsecond}0.373
& \cellcolor{heatsecond}0.318 \\
\bottomrule
\end{tabular}}
\caption{Quantitative comparison on out-of-domain datasets.
We evaluate against baselines on Mip-NeRF 360~\cite{barron2022mip} and
NeRFBusters~\cite{warburg2023nerfbusters} under \{3, 6, 9\} views.
$^{*}$ denotes results reproduced by us using their official implementation.
Colors from
{\setlength{\fboxsep}{1pt}\colorbox{heatfourth}{yellow}} to
{\setlength{\fboxsep}{1pt}\colorbox{heatbest}{red}}
indicate increasing performance.}
\label{tab:ood}
\end{table*}

\begin{figure}[t] 
\centering \includegraphics[width=1\linewidth]{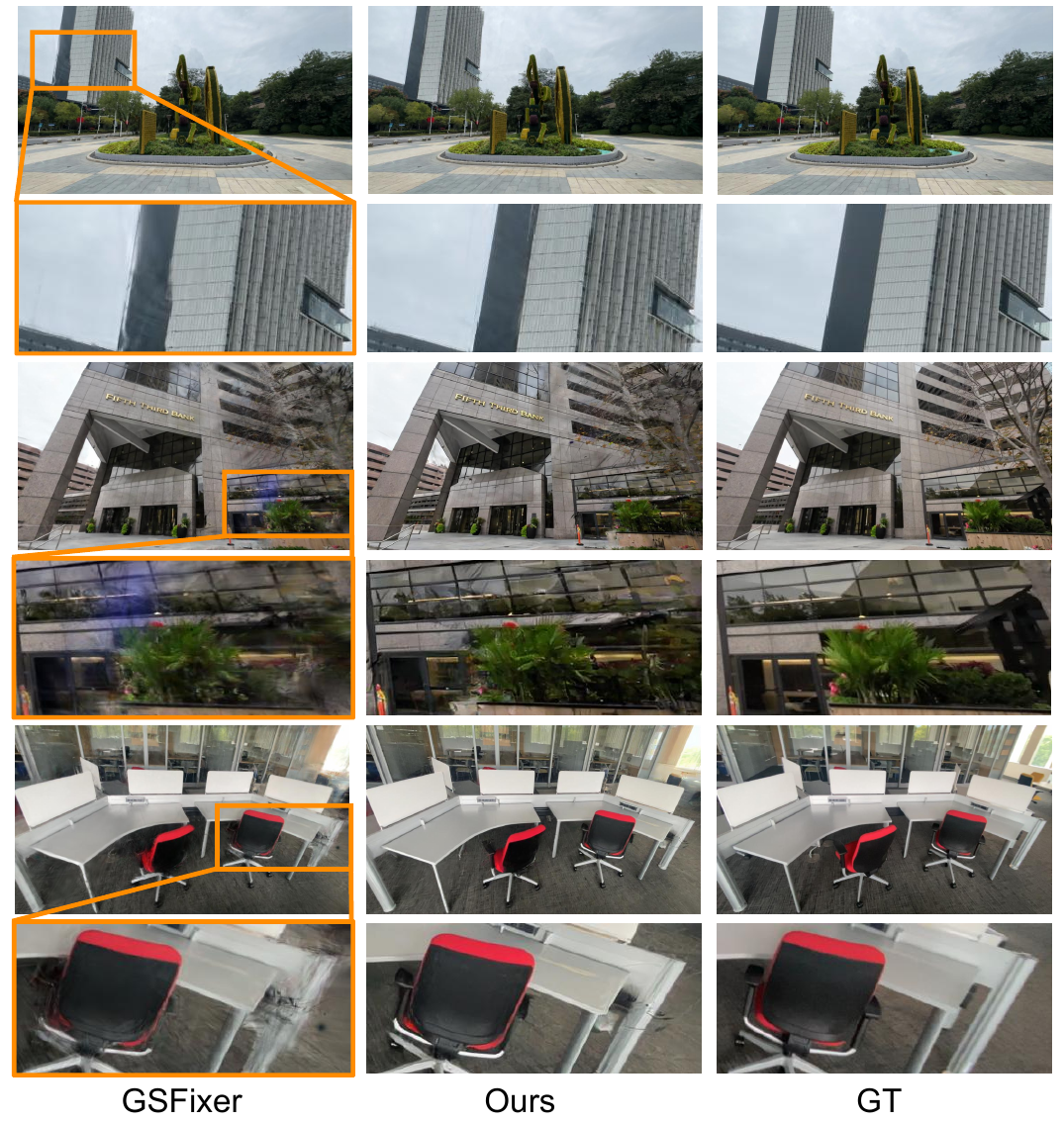} 
\caption{Qualitative comparison on DL3DV. GSFixer blurs reflective and
under-observed regions (zoomed), whereas TRACE-GS restores sharper
detail closer to the GT.}
\label{fig:dl3dv} 
\end{figure}

\noindent\textbf{Evaluation Datasets and Metrics.}
We evaluate on three benchmark datasets:
\textbf{1) DL3DV-Benchmark}~\cite{ling2024dl3dv},
\textbf{2) Mip-NeRF~360}~\cite{barron2022mip}, and
\textbf{3) NeRFBusters}~\cite{warburg2023nerfbusters}.
We report PSNR, SSIM, and LPIPS~\cite{zhang2018unreasonable}.

\subsection{Comparison with State-of-the-Art Methods}

\noindent\textbf{In-Domain Sparse-View Reconstruction.}
We first evaluate TRACE-GS on the DL3DV-Benchmark~\cite{ling2024dl3dv},
whose 28 scenes share the same data domain as our DL3DV training set
but are disjoint from the training scenes.
We compare against 3DGS~\cite{kerbl20233d},
Difix3D+~\cite{wu2025difix3d+}, GenFusion~\cite{wu2025genfusion}, and
GSFixer~\cite{yin2025gsfixer} under the 3-, 6-, and 9-view settings.

As shown in Table~\ref{tab:dl3dv}, TRACE-GS achieves the best PSNR
across all view settings, leads all metrics in the challenging 3-view
case, and remains best or second-best in SSIM as views increase.
Its PSNR gain over vanilla 3DGS grows from 2.02 to 2.35 and 3.20~dB as
views decrease from 9 to 6 and 3.
This trend is consistent with our geometry-asymmetric design: sparser
coverage makes the student's geometry less reliable, widening the
conditioning gap and thereby increasing the advantage of privileged
dense-view guidance; on-policy distillation applies this guidance along
the student's own rollout.

Fig.~\ref{fig:dl3dv} shows this on reflective and under-observed
regions: GSFixer blurs storefront reflections and chair contours,
whereas ours restores sharper, more faithful structure.

\begin{figure*}[!t]
\centering \includegraphics[width=1\linewidth]{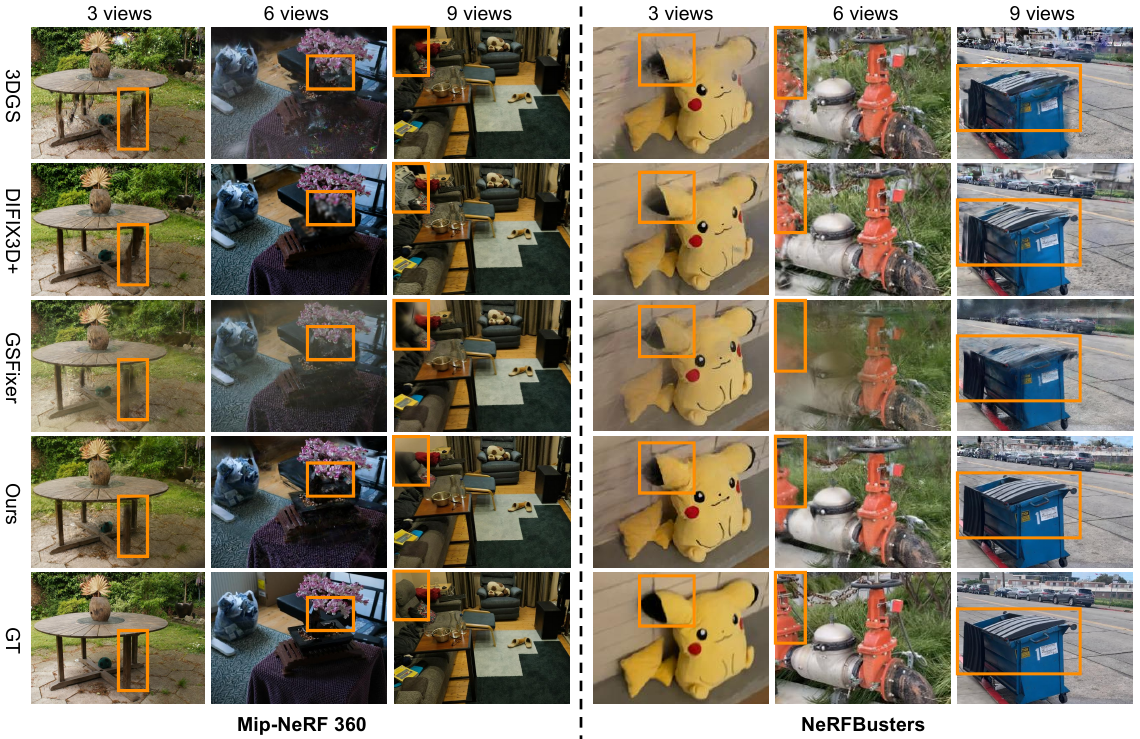} 
\caption{Qualitative comparison on out-of-domain sparse-view
reconstruction using 3, 6, and 9 views on Mip-NeRF~360~\cite{barron2022mip} (left) and
NeRFBusters~\cite{warburg2023nerfbusters} (right). \textcolor{orange}{Orange} boxes mark compared regions. Baseline
failures shift from broken coarse geometry at 3 views to oversmoothed
detail at denser settings, while TRACE-GS preserves both.}
\label{fig:ood}
\label{fig:ood} 
\end{figure*}

\noindent\textbf{Out-of-Domain Sparse-View Reconstruction.}
To assess generalization, we evaluate on Mip-NeRF
360~\cite{barron2022mip} and NeRFBusters~\cite{warburg2023nerfbusters},
which were unseen during training and differ from DL3DV in capture
style, scene scale, and camera trajectories.
\vspace{-1pt}

We compare against NeRF-based methods (Zip-NeRF~\cite{barron2023zip}
and ReconFusion~\cite{wu2024reconfusion}), splatting-based methods
(3DGS~\cite{kerbl20233d} and 2DGS~\cite{huang20242d}), and
diffusion-based 3DGS restoration methods
(Difix3D+~\cite{wu2025difix3d+}, GenFusion~\cite{wu2025genfusion}, and
GSFixer~\cite{yin2025gsfixer}).

As shown in Table~\ref{tab:ood}, TRACE-GS achieves the best PSNR and
SSIM at every sparsity level on both datasets and leads all three
metrics under 3 views as well as on Mip-NeRF~360 with 6 views.
Under 3 views, it outperforms GSFixer by 0.73~dB in PSNR on
Mip-NeRF~360 and 0.60~dB on NeRFBusters.

\vspace{-1pt}
Fig.~\ref{fig:ood} reveals the same progression on both datasets.
At 3 views, baselines lose coarse geometry: the table legs in
Mip-NeRF~360 fragment and Pikachu's ear structure in NeRFBusters
collapses, while ours preserves both.
As views increase, the bottleneck shifts to fine detail: GSFixer
oversmooths flower petals and sofa contours in Mip-NeRF~360 and surface
details on the hydrant and dumpster in NeRFBusters, whereas ours
preserves them.
Together, these results suggest that on-policy distillation transfers
training-time privileged guidance beyond the training distribution.

\subsection{Ablation Studies}

Our ablations evaluate the two requirements identified in Sec.~\ref{sec:intro}:
\textbf{(i)} reliable denoising directions and \textbf{(ii)} supervision at student-visited states.

\noindent\textbf{Effect of Privileged Geometry.}
Table~\ref{tab:geo} shows that increasing teacher views from 3 (B) to
12 (C), 18 (D), and all (F) monotonically improves every metric on both
benchmarks, reaching 19.15~dB on DL3DV and surpassing the no-teacher
baseline (A) by 0.51~dB.
Fig.~\ref{fig:policy} illustrates what this recovers: (A) is illegible,
whereas (D) restores the coarse layout of the text.
The non-privileged three-view teacher (B) falls below (A), indicating
that distillation alone provides no benefit without additional
geometric evidence.

\noindent\textbf{Effect of On-Policy Supervision.}
Variants (E) and (F) share the same all-view teacher and training
budget, differing only in the queried states: independently sampled in
(E) and student-visited in (F), thereby isolating the effect of the
off-policy state mismatch described in Sec.~\ref{sec:trajectory}.
On-policy supervision improves all metrics on both benchmarks.
The same comparison in Fig.~\ref{fig:policy} shows this on a text
region whose fine strokes are refined late in the reverse trajectory: (E)
bends strokes into letter-like but incorrect shapes, whereas (F)
resolves legible characters.

Fig.~\ref{fig:error} traces the same gap across denoising: the variants
remain close in early steps but separate in the later half, where the
off-policy variant plateaus while the on-policy variant continues to
improve.
Together, this structured distortion and late-emerging gap are
consistent with errors accumulating at student-visited states left
uncovered by independent-timestep supervision, rather than with a
capacity difference between the otherwise identical variants.
Even without $\mathcal{L}_\mathrm{ret}$, ($\mathrm{F}^{-}$) outperforms
(E) on all metrics, and retrieval alignment adds a further consistent
gain.

\begin{table}[t]
\centering
\setlength{\tabcolsep}{3pt}
\renewcommand{\arraystretch}{1.15}
\resizebox{\columnwidth}{!}{%
\begin{tabular}{llcccccccc} 
\toprule
    &               & Teacher & Priv. & \multicolumn{3}{c}{DL3DV} & \multicolumn{3}{c}{Mip-NeRF 360} \\ 
\cmidrule(lr){5-7}\cmidrule(lr){8-10}
    & Method        & views   & geom. & PSNR$\uparrow$ & SSIM$\uparrow$ & LPIPS$\downarrow$ & PSNR$\uparrow$ & SSIM$\uparrow$ & LPIPS$\downarrow$ \\ 
\midrule
(A) & Baseline   & –   & –      & 18.64 & 0.612 & 0.382 & 17.17 & 0.426 & 0.486 \\ 
\hdashline
\rule{0pt}{2.2ex}(B) & Distil.  & 3   & \xmark & 18.36 & 0.608 & 0.389 & 16.93 & 0.421 & 0.489 \\
(C) & Distil.  & 12   & \cmark & 18.80 & 0.616 & 0.379 & 17.26          & 0.434          & 0.480 \\
(D) & Distil.  & 18  & \cmark & 18.92 & 0.619 & 0.375 & 17.59 & 0.440 & 0.471 \\ 
\hdashline
\rule{0pt}{2.2ex}(E) & Off-policy & all & \cmark & 18.98 & 0.619 & 0.374 & 17.53 & 0.443 & 0.474 \\
(F$^{-}$) & w/o $\mathcal{L}_\mathrm{ret}$ & all & \cmark & 19.09 & 0.622 & 0.372 & 17.77 & 0.450 & 0.466 \\
(F) & \textbf{Ours} & all & \cmark & \textbf{19.15} & \textbf{0.624} & \textbf{0.369} & \textbf{17.84} & \textbf{0.452} & \textbf{0.465} \\
\bottomrule
\end{tabular}}
\caption{
Ablation of privileged geometry and on-policy supervision on
DL3DV-Benchmark (in-domain) and Mip-NeRF~360 (out-of-domain), averaged
over 3, 6, and 9 student input views.
(A) is the frozen pretrained restorer; (B)--(F) share the same warm-up
initialization and training budget.
Under on-policy supervision, (B)--(D) and (F) increase teacher views
from three to all; (B) uses no privileged geometry.
Variant (E) instead uses independently sampled states, and
($\mathrm{F}^{-}$) sets $\lambda_{\mathrm{ret}}=0$.
``All'' denotes all training views.
}
\label{tab:geo}
\end{table}


\begin{figure}[h]
\centering
\includegraphics[width=1\linewidth]{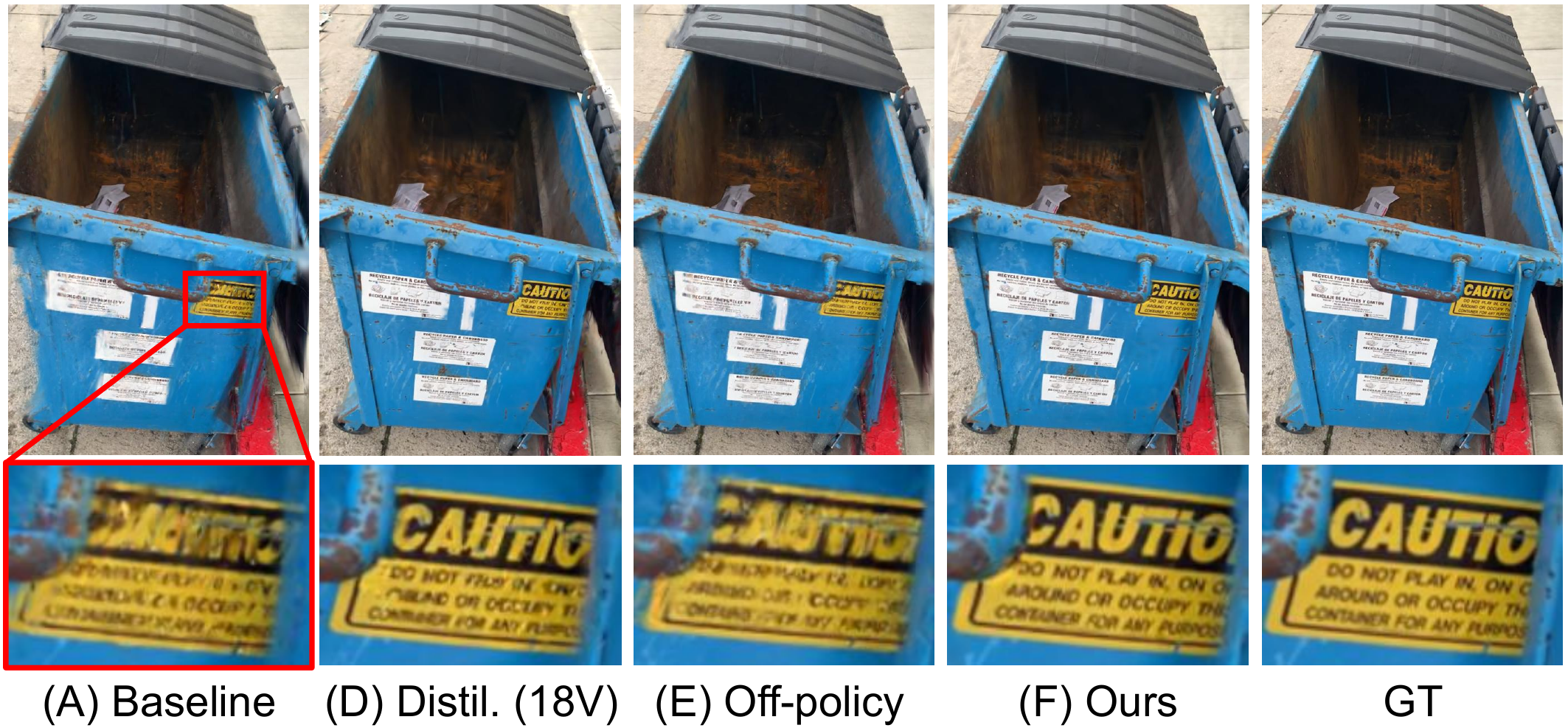}
\caption{Qualitative comparison on a zoomed text region.
Privileged geometry recovers coarse structure, while on-policy
supervision corrects fine strokes along the student rollout, yielding
text closest to GT.}
\label{fig:policy}
\end{figure}

\begin{figure}[h]
\centering
\includegraphics[height=0.15\textheight,width=\linewidth]{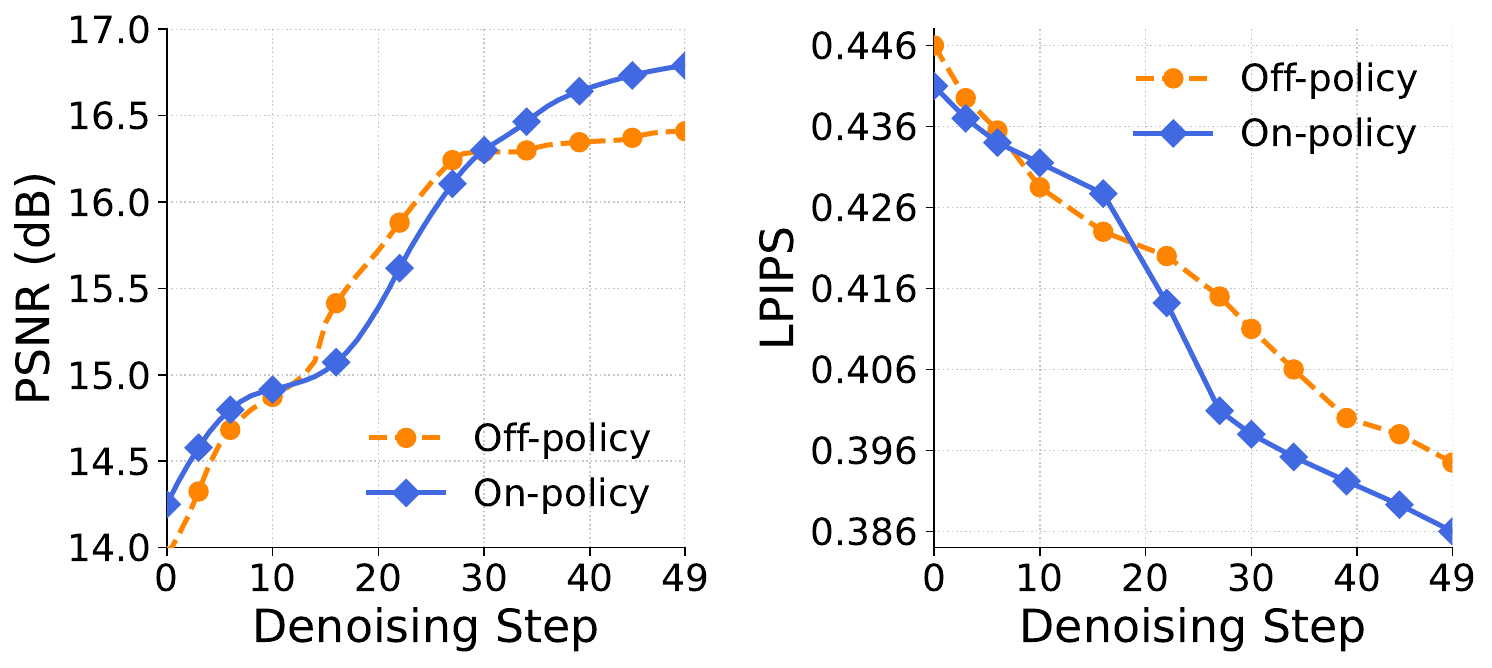}
\caption{Per-step restoration quality along the reverse diffusion
trajectory on held-out Mip-NeRF~360 clips with per-frame GT, averaged
across view settings before 3DGS refinement.
The variants remain close in early denoising steps but diverge later:
the off-policy variant plateaus, whereas the on-policy variant continues
to improve. This late-emerging gap is consistent with errors accumulating
at rollout states left uncovered by independent-state supervision.}
\label{fig:error}
\end{figure}

\section{Conclusion}
We introduce TRACE-GS, an on-policy trajectory distillation framework
that, to the best of our knowledge, is the first to derive on-policy supervision
from training-time privileged geometry for sparse-view 3DGS
restoration.
In its single supervision mechanism, this geometry defines the target,
and the student's own rollout determines where it is applied.
A dense-conditioned teacher is queried at these states to align
denoising directions and cross-view retrieval responses, without a
teacher rollout.
Only the sparse-conditioned student remains at deployment.
Across in- and out-of-domain benchmarks, TRACE-GS achieves consistent gains under 3, 6, and 9 views, most pronounced under severe sparsity.

\noindent\textbf{Limitations and Future Work.}
Our formulation targets static scenes, as in prior sparse-view
restoration work, and extends naturally to dynamic capture. Since
privileged geometry is training-only, deployment needs neither
additional views nor teacher computation, making TRACE-GS a drop-in
refinement stage for sparse capture. Future work includes dynamic
scenes and uncertainty-aware supervision.

\bibliography{aaai2027}

\clearpage
\setcounter{page}{1}

\twocolumn[
\begin{center}
    {\Large\bfseries Supplementary Material}
    \vspace{1em}
\end{center}
]

\appendix

\begin{center}
  {\large\scshape Table of Contents}
\end{center}
\vspace{2pt}
\tocrule
\vspace{10pt}

\tocsec{supp:impl}{A}{Further Implementation Details}
  \tocsub{supp:arch}{A.1}{Backbone}
  \tocsub{supp:cond}{A.2}{Conditioning Pathways}
  \tocsub{supp:lora}{A.3}{Teacher and Student Adapters}
  \tocsub{supp:train}{A.4}{Training Details}
  \tocsub{supp:deploy}{A.5}{Deployment}

\tocsec{supp:protocol}{B}{Evaluation Protocol}

\tocsec{supp:cost}{C}{Computational Cost}

\tocsec{supp:results}{D}{Additional Visual Results}

\tocsec{supp:video}{E}{Supplementary Video}

\vspace{12pt}
\tocrule
\vspace{10pt}

\noindent
This supplement specifies the implementation of TRACE-GS beyond what
space permits in the main paper. Sec.~\ref{supp:impl} describes the
backbone, conditioning interfaces, LoRA adapters, two-stage training
schedule, and deployment-time refinement loop. Together, these details
make precise the claim of Sec.~3.2 that, at each paired comparison
state, the only role-dependent external condition is the rendered
sequence: the student receives $R_s$, whereas the teacher receives
$R_d$. Sec.~\ref{supp:protocol} states the evaluation protocol and the
sense in which our comparisons are matched.
Sec.~\ref{supp:cost} characterizes the computational cost of on-policy
supervision at training and deployment time.
Sec.~\ref{supp:results} reports additional visual results, and
Sec.~\ref{supp:video} describes the accompanying downloadable
supplementary video, which compares novel-view sequences along
continuous camera trajectories.

Unless stated otherwise, references to numbered equations, tables,
figures, and algorithms refer to the main paper. Notation follows the
main paper: lowercase subscripts index view sets and their renderings
($s$ for sparse and $d$ for dense), whereas uppercase subscripts index
the two conditioning roles ($S$ for the sparse-conditioned student and
$T$ for the dense-conditioned teacher). Thus, the student is
conditioned on $R_s$, while the teacher is conditioned on $R_d$.

\section{Further Implementation Details}
\label{supp:impl}

\subsection{Backbone}
\label{supp:arch}

The frozen restoration backbone $\theta$ is a latent video DiT built
on CogVideoX-5B-I2V~\cite{yang2024cogvideox} and augmented with
reference cross-attention. TRACE-GS retains this restoration
architecture and introduces no additional deployment-time module; its
changes are confined to training-time supervision, state sampling, and
the teacher--student formulation. All pretrained base weights,
including the reference-conditioning modules, remain frozen throughout
training. Only rank-$32$ LoRA adapters are trainable, and during
Stage~2 only the student adapter $\phi_S$ is optimized
(Sec.~\ref{supp:lora}).

The reference-guided restoration initialization and conditioning
interface follow GSFixer~\cite{yin2025gsfixer}. The backbone contains
$42$ transformer blocks with hidden width $d=3072$, implemented using
$48$ attention heads with per-head dimension $d_h=64$, and operates in
the latent space of a frozen causal 3D VAE. The principal sublayers of
each transformer block are:
\begin{enumerate}[label=(\roman*),leftmargin=*,itemsep=2pt]
\item joint 3D self-attention over the concatenated video and text
tokens;
\item reference cross-attention, whose queries are the video-token
hidden states and whose keys and values are derived from the
timestep-invariant reference tokens $\mathbf{T}_{\mathrm{ref}}$
(Sec.~\ref{supp:cond});
\item a feed-forward network with expansion factor $4$
($d_{\mathrm{ff}}=12{,}288$).
\end{enumerate}
Timestep conditioning is injected through expert adaptive layer
normalization driven by a $512$-dimensional timestep embedding.

Each clip contains $49$ frames at $480\times720$ resolution. The frozen
VAE compresses the spatial dimensions by a factor of $8$. With a
temporal compression factor of $4$ and causal boundary handling, it
maps the temporal dimension to
$1+(49-1)/4=13$ latent frames. The resulting
$13\times60\times90$ latent grid is patchified using $2\times2$
spatial patches, yielding
$13\times30\times45=17{,}550$ video tokens per clip.

\subsection{Conditioning Pathways}
\label{supp:cond}

In addition to the denoising state and timestep, the backbone receives
three content-conditioning inputs: the rendered sequence, the
reference tokens, and the text tokens. We describe each in turn and
then establish that only the rendered-sequence input is role-dependent.

\paragraph{Rendered-sequence condition.}
For $b\in\{s,d\}$, the frozen VAE encodes the 3DGS-rendered sequence
$R_b$ into a latent condition $\operatorname{Enc}_{\mathrm{VAE}}(R_b)$
on the same spatiotemporal grid as the denoising state. This is the VAE
encoding of the rendered pixels and is distinct from the dashed
\emph{encodes} association in Eq.~(1), which denotes the geometry $c_b$
implicitly carried by $R_b$. Let $\xi$ denote the denoising state
supplied to both roles: $\xi=z_t$ at a forward-noised Stage-1 state and
$\xi=z^S_{t'_i}$ at a student-visited on-policy state. The denoising
and rendered-sequence latents are concatenated along the channel
dimension and passed through the shared patch-embedding
projection~\cite{yin2025gsfixer}:
\begin{equation}
X_b =
\operatorname{PatchEmbed}
\left(
    \operatorname{Concat}_{\mathrm{ch}}
    \left[
        \xi,\operatorname{Enc}_{\mathrm{VAE}}(R_b)
    \right]
\right).
\label{eq:supp-render-cond}
\end{equation}
The resulting tokens $X_b$ form the video-token input to the DiT, where
they are concatenated with the text tokens for joint self-attention.
The VAE encoder and patch-embedding projection belong to the frozen
base model and are identical for the two roles. Because $R_b$ is independent of
the denoising timestep, $\operatorname{Enc}_{\mathrm{VAE}}(R_b)$ is
computed once per clip and reused across all queried denoising states.

The geometry asymmetry is carried implicitly by these rendered pixels.
$R_s$ and $R_d$ originate from 3DGS representations constrained by
different numbers of input views, so the geometry they carry, $c_s$ and
$c_d$ in the notation of Eq.~(1), differs in reliability. This
distinction is relative rather than absolute: $c_d$ need not represent
ground-truth geometry; TRACE-GS only relies on it being, on average,
better constrained than $c_s$, so that the dense-conditioned prediction
provides a more reliable supervisory direction. We introduce no
role-specific depth, normal, point-cloud, camera-parameter, or other
explicit geometry channel.

\paragraph{Reference tokens.}
Each clip is anchored by two observed reference views
$\{I^{\mathrm{ref}}_1,I^{\mathrm{ref}}_2\}\subset\mathcal{V}_s$,
which define the beginning and end of the clip
trajectory~\cite{yin2025gsfixer}. Intermediate target views are sampled
between these anchors, yielding a $49$-frame clip with two reference
views and $47$ target views.

The frozen DINOv2 and VGGT encoders extract semantic and geometric
features from the reference views, respectively. DINOv2 processes each
view independently, whereas VGGT jointly processes the reference pair
to capture their multi-view geometric relationship. For $r\in\{1,2\}$,
the two feature types are projected to the DiT hidden dimension and
fused by addition, and the reference tokens are the concatenation over
both views:
\begin{equation}
\begin{aligned}
\mathbf{T}^{(r)}_{\mathrm{ref}}
={}&
\operatorname{Proj}_{\mathrm{2D}}
\left(
    \operatorname{Enc}_{\mathrm{DINO}}
    (I^{\mathrm{ref}}_r)
\right)
\\
&+
\operatorname{Proj}_{\mathrm{3D}}
\left(
    \left[
        \operatorname{Enc}_{\mathrm{VGGT}}
        (I^{\mathrm{ref}}_1,I^{\mathrm{ref}}_2)
    \right]_r
\right),
\end{aligned}
\label{eq:supp-ref-fusion}
\end{equation}
\begin{equation}
\mathbf{T}_{\mathrm{ref}}
=
\operatorname{Concat}_{\mathrm{tok}}
\left[
    \mathbf{T}^{(1)}_{\mathrm{ref}},
    \mathbf{T}^{(2)}_{\mathrm{ref}}
\right],
\label{eq:supp-ref-tokens}
\end{equation}
where $[\cdot]_r$ selects the tokens associated with the $r$-th
reference view. The reference tokens are computed once per clip and
reused at every denoising timestep. In each DiT block, the video-token
hidden states provide the queries, while $\mathbf{T}_{\mathrm{ref}}$
provides the keys and values for the reference cross-attention.

Both reference views are selected from
$\mathcal{V}_s\subset\mathcal{V}_d$. The teacher and student therefore
receive exactly the same reference images and the same
$\mathbf{T}_{\mathrm{ref}}$, although the additional views in
$\mathcal{V}_d\setminus\mathcal{V}_s$ are used to construct the
teacher's rendered condition $R_d$. The reference tokens thus carry
shared semantic and geometric evidence, but that evidence is identical
for the two roles and does not contribute to the teacher--student
asymmetry.

\paragraph{Text tokens.}
We retain the T5 text-conditioning pathway of CogVideoX. For each
paired clip, a frozen BLIP2 captioner generates a sample-specific
caption from the first frame of the sparse artifact-rendered sequence
$R_s$. We append the fixed quality suffix
``\texttt{. The video is of high quality, and the view is very clear.
High quality, masterpiece, best quality, highres, ultra-detailed,
fantastic.}'' and encode the resulting prompt with the frozen T5 text
encoder, following the restoration prompt setup of
GSFixer~\cite{yin2025gsfixer}. The positive text embedding is computed
once per clip and reused across denoising states.

To support classifier-free guidance, we apply caption dropout during
training: with probability $0.1$, the positive text embedding is
replaced by an all-zero embedding rather than by the T5 encoding of an
empty string. The dropout mask is sampled once per paired sample and
shared between the two roles. Consequently, the teacher and student
receive identical text tokens, whether the text condition is retained
or dropped. Text conditioning is therefore sample-dependent but
role-independent.

\paragraph{Localization of the role-dependent signal.}
As established above, the reference and text tokens are identical for
the two roles, as are the denoising state and timestep at each
comparison state. Their rendered sequences $R_s$ and $R_d$ are
generated along the same target trajectory $\mathcal{P}$, so
corresponding frames are pose-aligned, although $\mathcal{P}$ is not
supplied as an explicit model input. The only role-dependent external
input is therefore the rendered-sequence latent
$\operatorname{Enc}_{\mathrm{VAE}}(R_b)$ in
Eq.~\eqref{eq:supp-render-cond}: the privileged geometry $c_d$ affects
the teacher only through the rendered pixels of $R_d$.

The retrieval alignment in Eq.~(7) therefore compares how two
capacity-matched paths, conditioned on $R_s$ and $R_d$, retrieve the
same reference evidence; its discrepancy cannot be attributed to
different reference inputs or model-capacity asymmetry.

\begin{table*}[t]
\centering
\small
\begin{tabular*}{\textwidth}{@{\extracolsep{\fill}}llc@{}}
\toprule
Module & Adapted projections & LoRA \\
\midrule
Joint 3D self-attention
& \texttt{to\_q,to\_k,to\_v,to\_out.0}
& \checkmark \\
Reference cross-attention
& \texttt{to\_q,to\_k,to\_v,to\_out.0}
& \checkmark \\
Feed-forward network
& ---
& \ding{55} \\
\midrule
Patch embedding and input projection
& ---
& \ding{55} \\
Expert adaptive layer normalization
& ---
& \ding{55} \\
Normalization layers and biases
& ---
& \ding{55} \\
Final output projection
& ---
& \ding{55} \\
VAE, DINOv2, VGGT, and T5 encoders
& ---
& \ding{55} \\
\bottomrule
\end{tabular*}
\caption{
LoRA insertion points, identical for the teacher and student adapters.
Adapters are instantiated in both attention modules of all $42$
transformer blocks, while the feed-forward and conditioning components
remain unadapted.
}
\label{tab:supp-lora-modules}
\end{table*}

\subsection{Teacher and Student Adapters}
\label{supp:lora}

The teacher and student are matched exactly in architecture and
capacity. Both roles use rank-$32$ LoRA adapters attached to the same
frozen base model, with identical rank, scaling, and target modules.
For each Stage-2 run, the teacher and student adapters are cloned from
the same aligned Stage-1 initialization. During Stage~2, their
parameter values subsequently diverge, since $\phi_T$ is frozen while
$\phi_S$ is optimized.

\paragraph{Low-rank update.}
For a linear layer
$W\in\mathbb{R}^{d_{\mathrm{out}}\times d_{\mathrm{in}}}$ of the frozen
base model, the adapted weight is
\begin{equation}
W'
=
W+\frac{\alpha}{r}BA,
\qquad
A\in\mathbb{R}^{r\times d_{\mathrm{in}}},
\quad
B\in\mathbb{R}^{d_{\mathrm{out}}\times r},
\label{eq:supp-lora}
\end{equation}
with rank $r=32$ and scaling $\alpha=32$. Hence $\alpha/r=1$ and
$W'=W+BA$. LoRA dropout is disabled. We use the default initialization
of our LoRA implementation, with the second factor $B$ initialized to
zero. Consequently, $BA=\mathbf{0}$ at initialization, and attaching
an adapter does not change the function computed by the pretrained
model. Before Stage~1, the adapter-augmented model is therefore
functionally identical to the frozen pretrained restorer evaluated as
variant~(A) of Table~3. Stage~1 then produces the aligned
initialization from which the Stage-2 teacher and student adapters are
cloned; no separately pretrained teacher or student model is used.

\paragraph{Insertion points.}
Table~\ref{tab:supp-lora-modules} lists the adapted projections.
Adapters are instantiated in both attention modules of all $42$
transformer blocks, giving
$42\times2\times4=336$ adapted linear layers and $672$ low-rank
factors. After the Stage-1 adapter is cloned, the LoRA updates attached
to \texttt{to\_q}, \texttt{to\_k}, and \texttt{to\_v} of the reference
cross-attention make $W^{b}_{Q}$, $W^{b}_{K}$, and $W^{b}_{V}$ in
Eq.~(7) path-specific during Stage~2. The two roles retrieve from the
same reference tokens $\mathbf{T}_{\mathrm{ref}}$ through
capacity-matched but separately parameterized projections, whose
responses are aligned by $\mathcal{L}_{\mathrm{ret}}$.

\paragraph{Parameter budget.}
Because the reference features are projected to the DiT hidden width
before reference cross-attention, every adapted projection is square,
with $d_{\mathrm{in}}=d_{\mathrm{out}}=d=3072$. Each projection
therefore contributes
$r(d_{\mathrm{in}}+d_{\mathrm{out}})=2rd$ parameters. With four adapted
projections in each of the two attention modules, one transformer block
contributes
\begin{equation}
n_{\mathrm{blk}}
=
\underbrace{8rd}_{\text{joint self-attention}}
+
\underbrace{8rd}_{\text{reference cross-attention}}
=
16rd .
\label{eq:supp-lora-block-params}
\end{equation}
One adapter consequently contains
$42\times16\times32\times3072=66{,}060{,}288$ parameters
($66.06$\,M), less than $1.5\%$ of the frozen DiT backbone.

\paragraph{Adapter pairing.}
Stage~1 trains a single adapter $\phi$ shared by both roles. For each
Stage-2 run, the resulting aligned initialization is loaded once and
used to form
\begin{equation}
\phi_T
\leftarrow
\operatorname{Freeze}
\bigl(\operatorname{Clone}(\phi)\bigr),
\qquad
\phi_S
\leftarrow
\operatorname{Clone}(\phi),
\label{eq:supp-adapter-clone}
\end{equation}
as summarized in Alg.~1. The two adapters are deep copies of the same
tensors: no re-initialization occurs at cloning, no gradient path
connects them thereafter, and $\phi_T$ is evaluated without gradient
tracking. Because they share the same rank, scaling, target modules,
and frozen backbone, the pair remains capacity-matched throughout
Stage~2. Differences among the variants in Table~3 therefore cannot be
attributed to teacher--student model-size asymmetry.

The two roles share all frozen base-model weights and differ in
parameter storage only through their adapter sets. We therefore retain
a single copy of the frozen backbone and activate $\phi_T$ or $\phi_S$
for the corresponding forward pass. Relative to student-only training,
the incremental cost is one frozen $66.06$\,M-parameter adapter in
storage and one no-gradient teacher forward at each queried state
(Sec.~\ref{supp:cost}).

\subsection{Training Details}
\label{supp:train}

\paragraph{Pseudo-target generation.}
The training targets $\widehat{I}_T$ are generated once, offline,
before either training stage. For each paired clip, a frozen pretrained
GSFixer restorer is applied to the dense-view rendering $R_d$ to
produce $\widehat{I}_T$. This generator uses the same frozen backbone
weights $\theta$ as the teacher and student, together with the fixed
pretrained GSFixer adapter $\phi_{\mathrm{pre}}$. Because it is
conditioned on $R_d$, its outputs are derived from renderings whose
geometry is typically better constrained than that of $R_s$.

The generator adapter $\phi_{\mathrm{pre}}$ is separate from the
Stage-1 adapter $\phi$ and from the Stage-2 adapters $\phi_T$ and
$\phi_S$. The pseudo-target generator should therefore not be confused
with the privileged teacher used during Stage~2: the former produces
$\widehat{I}_T$ offline for the flow-matching supervision, whereas the
latter is queried at student-visited states to provide velocity and
retrieval targets. The offline generator is never evaluated during
Stage~2.

\paragraph{Two-stage schedule.}
Training proceeds in two stages that are run as separate jobs rather
than as two phases of a single optimizer trajectory. Stage~1 runs for
$500$ iterations and optimizes
\begin{equation}
\mathcal{L}_{\mathrm{warm}}
=
\mathcal{L}^{S}_{\mathrm{fm}}
+
\mathcal{L}^{T}_{\mathrm{fm}}
+
\lambda_{\mathrm{align}}
\mathcal{L}_{\mathrm{align}},
\label{eq:supp-stage1}
\end{equation}
at states obtained by forward-noising the latent encoding of
$\widehat{I}_T$. This objective combines flow-matching supervision on
$\widehat{I}_T$ for both roles with the privileged-direction alignment
of Sec.~3.3. The flow-matching terms anchor both predictions to the
pseudo-target velocity and discourage a conditioning-invariant
shortcut in which the shared adapter ignores the rendered-sequence
condition.

A single adapter $\phi$ is shared by the two roles throughout
Stage~1. Although $v_T$ is detached within
$\mathcal{L}_{\mathrm{align}}$, $\phi$ is updated by the complete
Stage-1 objective, including $\mathcal{L}^{T}_{\mathrm{fm}}$.
Consequently, the dense-conditioned prediction $v_T$ is recomputed at
every Stage-1 optimization update and serves as an online alignment
target, in contrast to the frozen teacher used in Stage~2. The
resulting adapter is saved as the Stage-1 initialization.

Stage~2 loads this checkpoint and forms the frozen teacher $\phi_T$
and trainable student $\phi_S$ according to
Eq.~\eqref{eq:supp-adapter-clone}. It then optimizes
\begin{equation}
\mathcal{L}
=
\mathcal{L}_{\mathrm{traj}}
+
\lambda_{\mathrm{ret}}
\mathcal{L}_{\mathrm{ret}}
\label{eq:supp-stage2}
\end{equation}
for $2{,}000$ iterations. In the full TRACE-GS configuration, this
objective is evaluated along student rollouts; the off-policy ablation
instead evaluates the teacher--student objective at independently
forward-noised states. The first $100$ Stage-2 iterations use linear
learning-rate warm-up. This optimizer warm-up is distinct from
Stage~1, whose objective is denoted
$\mathcal{L}_{\mathrm{warm}}$ in
Eq.~\eqref{eq:supp-stage1} for consistency with the main paper.

Variants~(B)--(F) of Table~3 use the same Stage-1 initialization and
the same number of Stage-2 optimization iterations. Stage~1 is held
fixed across these variants; the teacher-view and privileged-geometry
entries in Table~3 refer to the Stage-2 teacher configuration. The
variants then differ in the number of Stage-2 teacher views and the
states at which the teacher is queried. Variant~$(F^{-})$ additionally
sets $\lambda_{\mathrm{ret}}=0$.

Training is iteration-based rather than epoch-based: at each
optimization step, a mini-batch is drawn from the precomputed paired
clip pool. The reported iteration counts therefore denote optimizer
updates rather than complete passes over all training clips.

\paragraph{Detached rollout and gradient flow.}
In the full on-policy configuration, each Stage-2 iteration constructs
an $N=10$-step student rollout on a time grid
$\{t'_i\}_{i=0}^{N}$ resampled from the $M=50$-step deployment
schedule. At each visited state $z^{S}_{t'_i}$, the student forward pass
is evaluated with gradient tracking. The frozen teacher is queried
without gradient tracking at the same state and timestep, providing
the corresponding velocity and retrieval targets without performing a
separate teacher rollout.

After computing the local losses at step $i$, the student state is
advanced according to
\begin{equation}
\begin{aligned}
u_i^S
&=
v_{\theta,\phi_S}
\left(
    z^{S}_{t'_i},
    t'_i,
    R_s
\right),
\\
z^{S}_{t'_{i-1}}
&=
\operatorname{sg}
\left[
    \mathcal{T}_i
    \left(
        z^{S}_{t'_i},
        u_i^S
    \right)
\right],
\end{aligned}
\label{eq:supp-detached-transition}
\end{equation}
where $\operatorname{sg}$ denotes stop-gradient. The losses from all
$N$ visited states are averaged, followed by one backward pass and one
optimizer update per rollout. Gradients from every local loss
contribute to updating the student adapter $\phi_S$, but no gradient
propagates through a scheduler transition or between successive
rollout steps.

This avoids full backpropagation through the rollout while still
supervising states the student actually visits. Because the student
computation graphs for all $N$ local losses are retained until the
shared backward pass, student-side activation memory increases with the
rollout length, whereas teacher activations are not retained for
backward (Sec.~\ref{supp:cost}).

\paragraph{Retrieval-response read-out.}
Equation~(7) omits the transformer-block index for clarity. In all
experiments, $F_i^b$ is read from the reference cross-attention module
of the final transformer block ($m^\star=42$). This location is fixed
a priori and is not selected through layer search. We use the final
block because its reference retrieval response is closest to the
predicted denoising velocity, while its query hidden states already
incorporate the processing of all preceding blocks.

At student-visited state $i$, the video-token hidden states of the
final block provide the queries, while the shared reference tokens
provide the keys and values:
\begin{equation}
\begin{aligned}
Q_i^b
&=
H_{i,m^\star}^b
W_{Q,m^\star}^b,
\\
K^b
&=
\mathbf{T}_{\mathrm{ref}}
W_{K,m^\star}^b,
\qquad
V^b
=
\mathbf{T}_{\mathrm{ref}}
W_{V,m^\star}^b,
\end{aligned}
\label{eq:supp-retrieval-qkv}
\end{equation}
where $b\in\{S,T\}$ denotes the sparse-conditioned student or the
dense-conditioned teacher. The reference tokens are identical for the
two roles, whereas the projections are separately parameterized after
the Stage-1 adapter is cloned.

We retain the value-weighted response of all $48$ attention heads
before the output projection \texttt{to\_out.0}:
\begin{equation}
F_i^b
=
\operatorname{Concat}_{h=1}^{48}
\left[
\operatorname{Softmax}
\left(
\frac{
Q_{i,h}^b
(K_h^b)^{\top}
}{
\sqrt{d_h}
}
\right)
V_h^b
\right],
\label{eq:supp-retrieval-readout}
\end{equation}
where $d_h=64$ is the per-head dimension, so that
$F_i^b\in\mathbb{R}^{n_{\mathrm{vid}}\times d}$ with $d=3072$. For
each video token $q$, we normalize its response independently along
the channel dimension:
\begin{equation}
\bar{F}_{i,q}^b
=
\frac{
F_{i,q}^b
}{
\lVert F_{i,q}^b\rVert_2+\epsilon
},
\label{eq:supp-retrieval-normalization}
\end{equation}
where $\epsilon$ is a small constant for numerical stability. The
retrieval-alignment loss is
\begin{equation}
\mathcal{L}_{\mathrm{ret}}
=
\mathbb{E}
\left[
\frac{1}{N}
\sum_{i=1}^{N}
\operatorname{MSE}
\left(
\bar{F}_i^S,
\operatorname{sg}
\left[
\bar{F}_i^T
\right]
\right)
\right],
\label{eq:supp-retrieval-loss}
\end{equation}
where $\operatorname{MSE}$ uses mean reduction over the batch,
video-token, and channel dimensions. The stop-gradient on the teacher
response ensures that this term updates only $\phi_S$. We introduce no
layer subset, head selection, or cross-layer weighting, so
$\lambda_{\mathrm{ret}}$ is the only hyperparameter controlling this
term.

\subsection{Deployment}
\label{supp:deploy}

\paragraph{Student-only inference.}
Of the two adapters, only $\phi_S$ is retained after training.
Deployment uses the same backbone, conditioning interface, denoising
schedule, and number of diffusion evaluations as a standard GSFixer
restoration pass, and differs only in the values of a same-sized LoRA
adapter. Neither the teacher adapter $\phi_T$, the dense-view
reconstruction, nor the additional views
$\mathcal{V}_d\setminus\mathcal{V}_s$ is used at any deployment stage,
and no teacher query or additional diffusion rollout is performed.
Privileged geometry therefore introduces no deployment-time overhead
(Sec.~\ref{supp:cost}).

\paragraph{Iterative refinement.}
We index deployment quantities by the repair round
$r\in\{1,\ldots,N_r\}$; the main paper suppresses this index. Given a
scene reconstructed from $K$ sparse views, deployment alternates
student restoration and 3DGS optimization for $N_r=3$ repair rounds. The
representation is optimized for $J=30{,}000$ iterations in total, with
repair invoked at
\begin{equation}
\mathcal{J}_{\mathrm{repair}}
=
\{7{,}000,\;17{,}000,\;27{,}000\}.
\label{eq:supp-repair-steps}
\end{equation}
The first $7{,}000$ iterations constitute a sparse-only cold start,
during which the representation is optimized exclusively from the
captured views $\mathcal{V}_s$. The three repair calls are followed by
$10{,}000$, $10{,}000$, and $3{,}000$ additional 3DGS optimization
iterations, respectively.

At repair round $r$, a round-specific camera trajectory
$\mathcal{P}^{(r)}=\{P^{(r)}_n\}_{n=1}^{L_r}$ is resampled rather than
reused. It comprises a $360^{\circ}$ elliptical path whose scale and
vertical variation depend on $r$, with a randomly sampled vertical
phase, together with bridge poses connecting the observed reference
views to the ellipse. We render the current representation
$\mathcal{G}^{(r)}$ at these poses, obtaining
\begin{equation}
\widetilde{R}^{(r)}
=
\left\{
\operatorname{Render}
\left(
\mathcal{G}^{(r)},P^{(r)}_n
\right)
\right\}_{n=1}^{L_r}.
\label{eq:supp-round-rendering}
\end{equation}
The rendered trajectory is partitioned into $49$-frame clips compatible
with the restoration backbone, and each clip is restored by the
deployed student under the full $M=50$-step reverse-diffusion schedule. 
Every restored frame retains the pose of its input rendering,
yielding the posed pseudo-observation set
\begin{equation}
\widehat{\mathcal{V}}^{(r)}_S
=
\left\{
\left(
\widehat{I}^{(r)}_{S,n},
P^{(r)}_n
\right)
\right\}_{n=1}^{L_r}.
\label{eq:supp-round-pseudo-set}
\end{equation}

Within each repair round, restored frames from all clips are pooled
into a single pseudo-observation set. Across rounds, the preceding set
is cleared and replaced by $\widehat{\mathcal{V}}^{(r)}_S$: later
rounds do not accumulate pseudo-observations from earlier rounds but
replace them with restorations conditioned on the progressively refined
representation. No uncertainty, perceptual-quality, or confidence-based
filtering is applied when constructing $\widehat{\mathcal{V}}^{(r)}_S$;
every restored frame generated in the current round is added as a posed
pseudo-observation.

The captured sparse views remain active throughout optimization. At
iteration $j$, the sparse-view reconstruction loss retains unit weight,
while the pseudo-observation reconstruction loss is modulated by a
global half-sinusoidal schedule:
\begin{equation}
\mathcal{L}_{\mathrm{3DGS}}(j)
=
\mathcal{L}_{\mathrm{sparse}}
+
w_{\mathrm{reg}}(j)\,
\mathcal{L}_{\mathrm{pseudo}},
\label{eq:supp-refinement-objective}
\end{equation}
where $\mathcal{L}_{\mathrm{sparse}}$ and
$\mathcal{L}_{\mathrm{pseudo}}$ denote the reconstruction losses used
by the GSFixer refinement pipeline, evaluated on the captured views and
the current pseudo-observations, respectively, and
\begin{equation}
w_{\mathrm{reg}}(j)
=
\begin{cases}
0,
& 0\le j<7{,}000,
\\[3pt]
\lambda_{\mathrm{reg}}
\sin
\left(
\pi
\displaystyle\frac{j-7{,}000}{J-7{,}000}
\right),
& 7{,}000\le j\le J,
\end{cases}
\label{eq:supp-pseudo-weight}
\end{equation}
with $\lambda_{\mathrm{reg}}=1$. The schedule is defined once over the
complete refinement interval and is not restarted after each repair
call. It vanishes at both endpoints and reaches its maximum value of
$1$ at the midpoint, so generated content is introduced gradually
rather than at full strength when it first becomes available, and is
annealed out before optimization terminates. In contrast, the
captured-view term retains unit weight at every iteration, keeping the
representation anchored to the original observations rather than
allowing refinement to be driven by generated content alone.

\section{Evaluation Protocol}
\label{supp:protocol}

\paragraph{Baseline parity.}
Unless marked with an asterisk, baseline results in Tables~1 and~2 are
taken from the common evaluation reported in
GSFixer~\cite{yin2025gsfixer}. We use the same benchmark scenes,
$\{3,6,9\}$ sparse input views, target views, $480\times720$
evaluation resolution, and metric implementation. For methods
operating in the restore--refine regime, we additionally use the 3DGS
refinement protocol described in Sec.~\ref{supp:deploy}, including
$30{,}000$ optimization iterations, $N_r=3$ repair rounds, the same
trajectory-generation procedure, and the same pseudo-observation
weighting schedule. At evaluation, TRACE-GS replaces only the
restoration model within this pipeline; the reconstruction and
evaluation protocols are otherwise unchanged.

Results marked with an asterisk are obtained by running the official
implementation of ReconFusion~\cite{wu2024reconfusion}. We evaluate it
using the same sparse input views, target views, evaluation resolution,
and metric implementation as the other methods, and provide no
additional test-time views.

TRACE-GS and GSFixer further share the same pretrained
CogVideoX-based restoration backbone, frozen VAE, tokenization,
$49$-frame clip construction, two-reference conditioning interface,
and reference-view selection rule. At deployment, they use the same
initial reconstruction protocol, trajectory-generation rule,
$50$-step diffusion schedule, and downstream refinement settings. What
is matched is the procedure rather than the realized intermediate
renderings: after the first repair round, each method conditions on its
own progressively refined representation. Within the restoration
model, TRACE-GS differs from GSFixer in its training supervision and
the resulting student adapter.

\paragraph{Metric computation.}
PSNR, SSIM, and LPIPS are computed between rendered test views and
ground-truth images at $480\times720$ resolution, with LPIPS evaluated
using the VGG backbone. We first average each metric over the test
views of a scene and then report the unweighted mean across scenes.
The same image range, evaluation resolution, and metric implementation
are used for all of our runs. Baseline values taken from
GSFixer~\cite{yin2025gsfixer} are reported unchanged; the asterisked
ReconFusion results are evaluated with the same implementation as our
own runs.

\section{Computational Cost}
\label{supp:cost}

\paragraph{Training.}
Stage~1 adds a one-time $500$-iteration alignment run; the dominant
diffusion-training cost arises from Stage~2. Each Stage-2 iteration
evaluates the student and teacher at $N=10$ student-visited states,
requiring $N$ gradient-tracked student forwards, $N$ no-gradient
teacher forwards, and the detached scheduler transitions that
construct the rollout. The teacher is queried at the student states
rather than rolled out separately. The $N$ local losses are averaged,
followed by one backward pass and one optimizer update. Accordingly,
the number of student and teacher model evaluations per iteration
scales linearly with the rollout length $N$.

The two roles share one copy of the frozen backbone and use separate
$66.06$\,M-parameter adapters (Sec.~\ref{supp:lora}). Only $\phi_S$
requires gradients and optimizer states, and teacher activations are
not retained for backward. Because all $N$ student losses are retained
until the shared backward pass, student-side activation memory grows
with the rollout length. The retrieval loss reuses responses produced
during these forwards and requires no additional backbone evaluation.

Training-data construction is a one-time cost. For each of the $112$
training scenes, we construct the required sparse- and dense-view 3DGS
representations, render paired trajectories to obtain approximately
$150$ paired clips per scene, and generate dense-conditioned
pseudo-targets offline. On four NVIDIA A100 GPUs ($80$\,GB each), this
preprocessing together with the $2{,}000$ Stage-2 iterations takes
approximately five days end to end.

\paragraph{Deployment.}
TRACE-GS has the same algorithmic deployment cost as GSFixer under the
shared restore--refine protocol of Sec.~\ref{supp:deploy}: it uses the
same backbone, $M=50$ diffusion steps, and $N_r=3$ repair rounds, with
only the adapter values changed. Deployment runs on one NVIDIA A100
GPU ($80$\,GB). The teacher, dense-view reconstruction, and additional
training views are not used, so privileged geometry introduces no
additional deployment-time model evaluation, diffusion step, or repair
round.

\section{Additional Visual Results}
\label{supp:results}

We provide additional qualitative comparisons on Mip-NeRF
360~\cite{barron2022mip} and
NeRFBusters~\cite{warburg2023nerfbusters}.
Figures~\ref{fig:supp_mip} and~\ref{fig:supp_bus} compare TRACE-GS
with vanilla 3DGS~\cite{kerbl20233d} and three diffusion-based
restoration baselines:
Difix3D+~\cite{wu2025difix3d+},
GenFusion~\cite{wu2025genfusion}, and
GSFixer~\cite{yin2025gsfixer}.
Ground-truth views are shown for reference, and the examples cover the
$3$-, $6$-, and $9$-view settings.

All methods use the same sparse input views, target views, evaluation
resolution, and metric protocol described in
Sec.~\ref{supp:protocol}. Methods operating in the restore--refine
regime additionally use the matched downstream 3DGS refinement
schedule. Two visual behaviors recur across the examples. Some methods
retain residual haze, floaters, or colored streaks in under-observed
regions, whereas others suppress these artifacts at the cost of
over-smoothing texture or removing thin structures. The following
examples illustrate these recurring behaviors.

\paragraph{Qualitative comparison on Mip-NeRF 360.}
In Fig.~\ref{fig:supp_mip}, vanilla 3DGS and Difix3D+ retain visible
artifacts in several under-observed regions: floaters and colored
streaks remain above the flower bed, while residual haze appears near
the top of the \emph{kitchen} view. GenFusion suppresses many of these
artifacts but produces smoother results; for example, the
\emph{garden} table is covered by a hazy appearance that weakens its
radial plank texture. GSFixer generally recovers the coarse scene
layout but loses fine texture in regions with limited input-view
coverage.

In \emph{garden}, GSFixer renders the lower-right lawn as a largely
smoothed green region, whereas TRACE-GS better retains both the
granular grass appearance and the radial pattern of the table surface.
In \emph{flowers}, TRACE-GS produces more clearly separated flower-bed
boundaries and background foliage, while the GSFixer result contains
radial smearing around the bed and a blurred tree line. In
\emph{room}, GenFusion and GSFixer blur parts of the sofa and side-table
region into the background, whereas TRACE-GS more clearly preserves
the door frame, piano-bench legs, and reflections on the wooden floor.
In \emph{kitchen}, the chair back behind the model is also more clearly
recovered by TRACE-GS.

These improvements occur primarily in fine structures and
under-observed regions, where the sparse-view reconstruction provides
less reliable conditioning. The qualitative behavior is consistent
with the per-step analysis in Fig.~6: the off-policy variant plateaus
during the later denoising steps, whereas the on-policy variant
continues to improve as finer structure is resolved. It is also
consistent with the quantitative trends reported in Table~2.

\paragraph{Qualitative comparison on NeRFBusters.}
Figure~\ref{fig:supp_bus} presents the corresponding comparison on
NeRFBusters~\cite{warburg2023nerfbusters}. Its casually captured scenes
contain thin structures, small objects, and reflective surfaces that
are particularly sensitive to reconstruction and restoration errors.
The same broad failure patterns remain visible. Vanilla 3DGS contains
substantial haze in \emph{roses} and \emph{century}, obscuring the
bouquet and nearby table regions, and produces strong smearing over the
objects in \emph{table}. GenFusion removes much of the haze but also
alters local structure: the agave leaves in \emph{century} become
fragmented, with visible smearing around the planter, while the
tabletop in \emph{table} loses much of its surface detail.

The differences are especially visible on thin structures and small
objects. In \emph{roses}, the small white object near the lower-right
part of the table is absent from the 3DGS and GenFusion results and is
only weakly defined by GSFixer, whereas TRACE-GS more clearly recovers
its shape. TRACE-GS also better preserves the tabletop reflections. In
the NeRFBusters \emph{flowers} scene, GSFixer smears the fine
baby's-breath stems and introduces a red patch in the upper-left region
that is absent from the ground truth; TRACE-GS preserves the stems
without this visible artifact. In \emph{picnic}, TRACE-GS retains both
the wooden-slat texture and the background visible through the gaps,
whereas GenFusion fills several gaps with smoothed content. In
\emph{table}, TRACE-GS produces sharper boundaries around the circular
table edge and the objects resting on it.

The fabricated color patch in \emph{flowers}, together with the loss
of thin structures in the other examples, illustrates that plausible
per-frame restoration does not necessarily preserve the evidence
needed for stable novel-view reconstruction. These observations are
consistent with the motivation of Sec.~3.4: supervising the student at
its own visited states using targets derived from better-constrained
geometry reduces, although does not eliminate, the restoration errors
that can accumulate along the denoising rollout.



\section{Supplementary Video}
\label{supp:video}

A supplementary video is included with the submission as a separate
downloadable media supplement. It presents novel-view sequences
rendered along continuous camera trajectories. Static figures do not
fully reveal cross-view consistency: individually plausible frames may
still exhibit flicker, floaters, or geometry that changes across
viewpoints, and these effects become more apparent under continuous
motion. It complements static comparisons by revealing view-dependent
instability. For each scene, the video compares standard
3DGS~\cite{kerbl20233d}, GSFixer~\cite{yin2025gsfixer}, and TRACE-GS
using the same camera trajectory and synchronized playback, enabling
direct comparison of temporal flicker, floaters, and cross-view
geometric stability.

\begin{figure*}[t]
\centering \includegraphics[height=0.8\textheight,width=1\linewidth]{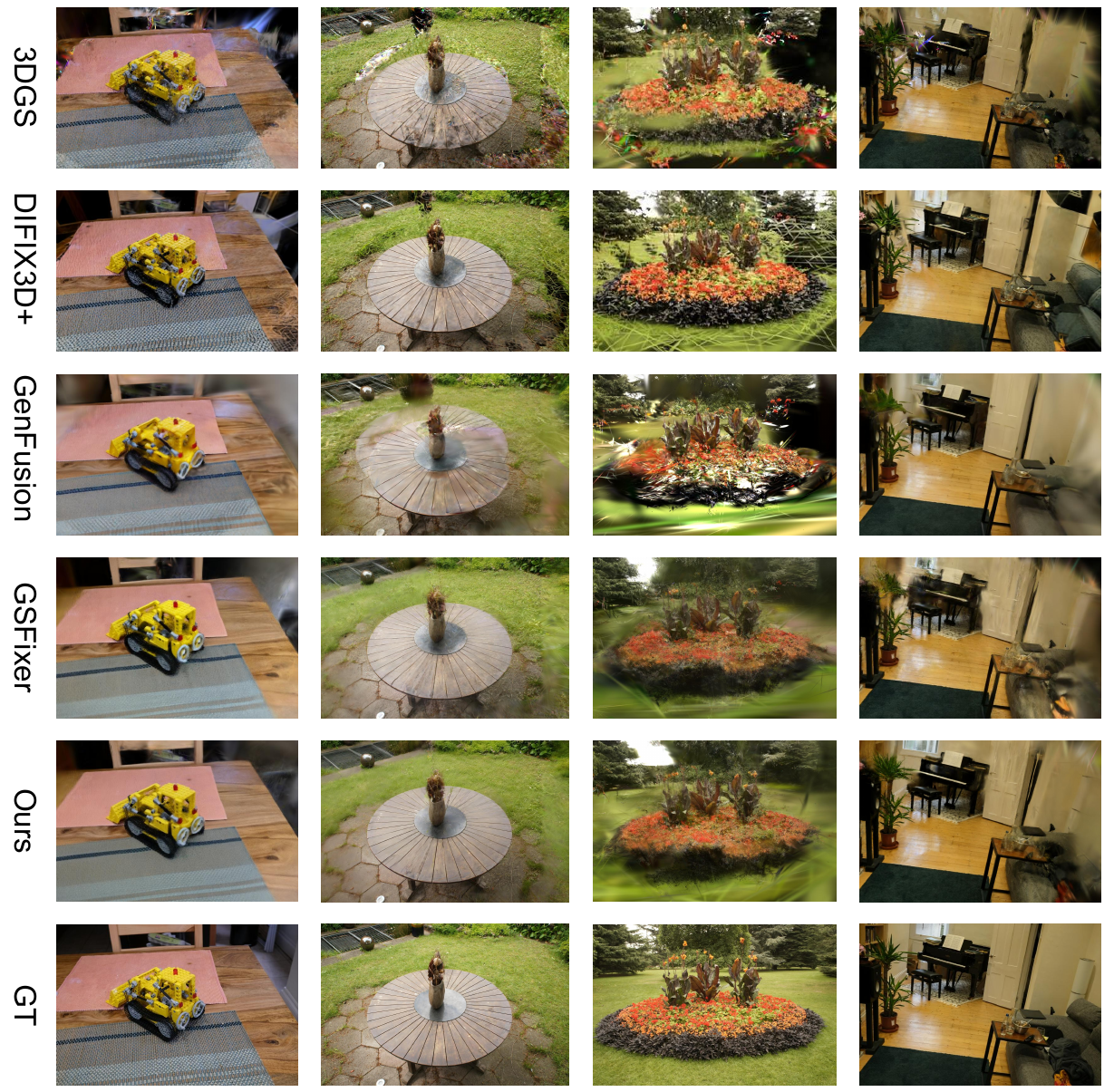} 
\caption{Qualitative comparison on out-of-domain sparse-view reconstruction on
Mip-NeRF 360~\cite{barron2022mip}. From left to right: \emph{kitchen},
\emph{garden}, \emph{flowers}, and \emph{room}. Examples are drawn from the
$3$-, $6$-, and $9$-view settings. Rows from top to bottom: 3DGS, Difix3D+,
GenFusion, GSFixer, TRACE-GS (ours), and ground truth. Baselines either leave
haze and floaters in under-observed regions or remove them at the cost of
over-smoothing; TRACE-GS reduces these artifacts while preserving more
high-frequency structure.}
\label{fig:supp_mip}
\end{figure*}

\newpage

\begin{figure*}[t]
\centering \includegraphics[height=1\textheight,width=0.82\linewidth]{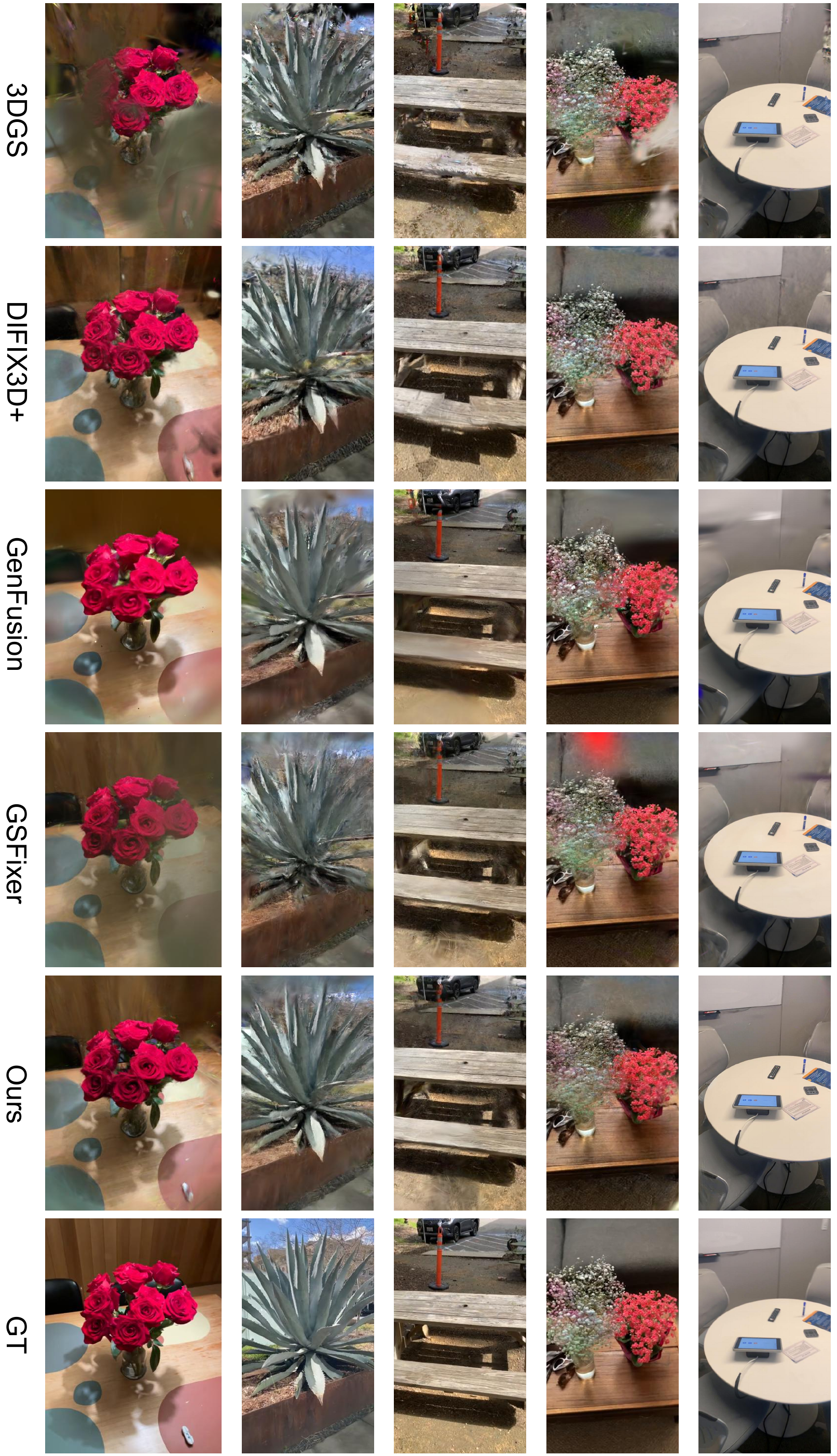} 
\caption{Qualitative comparison on out-of-domain sparse-view reconstruction on
NeRFBusters~\cite{warburg2023nerfbusters}, with examples drawn from the $3$-,
$6$-, and $9$-view settings. Columns: \emph{roses}, \emph{century},
\emph{picnic}, \emph{flowers}, \emph{table}.}
\label{fig:supp_bus}
\end{figure*}


\end{document}